\documentclass{article}

\usepackage{neurips_data_2024}

\usepackage[utf8]{inputenc} 
\usepackage[T1]{fontenc}    
\usepackage{hyperref}       
\usepackage{url}            
\usepackage{booktabs}       
\usepackage{amsfonts}       
\usepackage{nicefrac}       
\usepackage{microtype}      
\usepackage{xcolor}         
\usepackage{graphicx}
\usepackage{float} 
\usepackage{subfigure}
\usepackage{multirow}
\usepackage{tabularx}
\usepackage{mathrsfs}
\usepackage{adjustbox}  
\usepackage{caption}    
\usepackage{pifont}
\usepackage{ulem}
\usepackage{hyperref}
\usepackage{color}
\usepackage{amsmath}
\usepackage{chapterbib}
\usepackage{colortbl} 
\PassOptionsToPackage{table}{xcolor} 
\usepackage{xcolor}
\usepackage{threeparttable} 
\hypersetup{        
    colorlinks,
    linkcolor={red!50!black},
    citecolor={blue!50!black},
    urlcolor={blue!80!black}
}

\title{M4Fog: A Global Multi-Regional, Multi-Modal, and Multi-Stage Dataset for Marine Fog Detection and Forecasting to Bridge Ocean and Atmosphere}

\usepackage{authblk}

\author{
    \textbf{Mengqiu Xu}\textsuperscript{1}, \textbf{Ming Wu}\textsuperscript{1}, \textbf{Kaixin Chen}\textsuperscript{1}, \textbf{Yixiang Huang}\textsuperscript{1}, \textbf{Mingrui Xu}\textsuperscript{1}, \textbf{Yujia Yang}\textsuperscript{1}, \textbf{Yiqing Feng}\textsuperscript{1},\textbf{Yiying Guo}\textsuperscript{2}, \textbf{Bin Huang}\textsuperscript{2}, \textbf{Dongliang Chang}\textsuperscript{3}, \textbf{Zhenwei Shi}\textsuperscript{4}, \textbf{Chuang Zhang}\textsuperscript{1,5}, \textbf{Zhanyu Ma}\textsuperscript{1}, \textbf{Jun Guo}\textsuperscript{1} \\
    \textsuperscript{1} Beijing University of Posts and Telecommunications (BUPT), \textsuperscript{2} China National Meteorological Centre (NMC), \textsuperscript{3} Tsinghua University, \textsuperscript{4} Beihang University, \textsuperscript{5} Beijing Wuzi University
}

\begin{document}

\maketitle

\begin{abstract}

Marine fog poses a significant hazard to global shipping, necessitating effective detection and forecasting to reduce economic losses. In recent years, several machine learning (ML) methods have demonstrated superior detection accuracy compared to traditional meteorological methods. However, most of these works are developed on proprietary datasets, and the few publicly accessible datasets are often limited to simplistic toy scenarios for research purposes. To advance the field, we have collected nearly a decade's worth of multi-modal data related to continuous marine fog stages from four series of geostationary meteorological satellites, along with meteorological observations and numerical analysis, covering 15 marine regions globally where maritime fog frequently occurs. Through pixel-level manual annotation by meteorological experts, we present the most comprehensive marine fog detection and forecasting dataset to date, named \textbf{M4Fog}, to bridge ocean and atmosphere. The dataset comprises 68,000 "super data cubes" along four dimensions: elements, latitude, longitude and time, with a temporal resolution of half an hour and a spatial resolution of 1 kilometer. Considering practical applications, we have defined and explored three meaningful tracks with multi-metric evaluation systems: static or dynamic marine fog detection, and spatio-temporal forecasting for cloud images. Extensive benchmarking and experiments demonstrate the rationality and effectiveness of the construction concept for proposed M4Fog. The data and codes are available to whole researchers through cloud platforms \href{https://github.com/kaka0910/M4Fog}{\includegraphics[height=7pt]{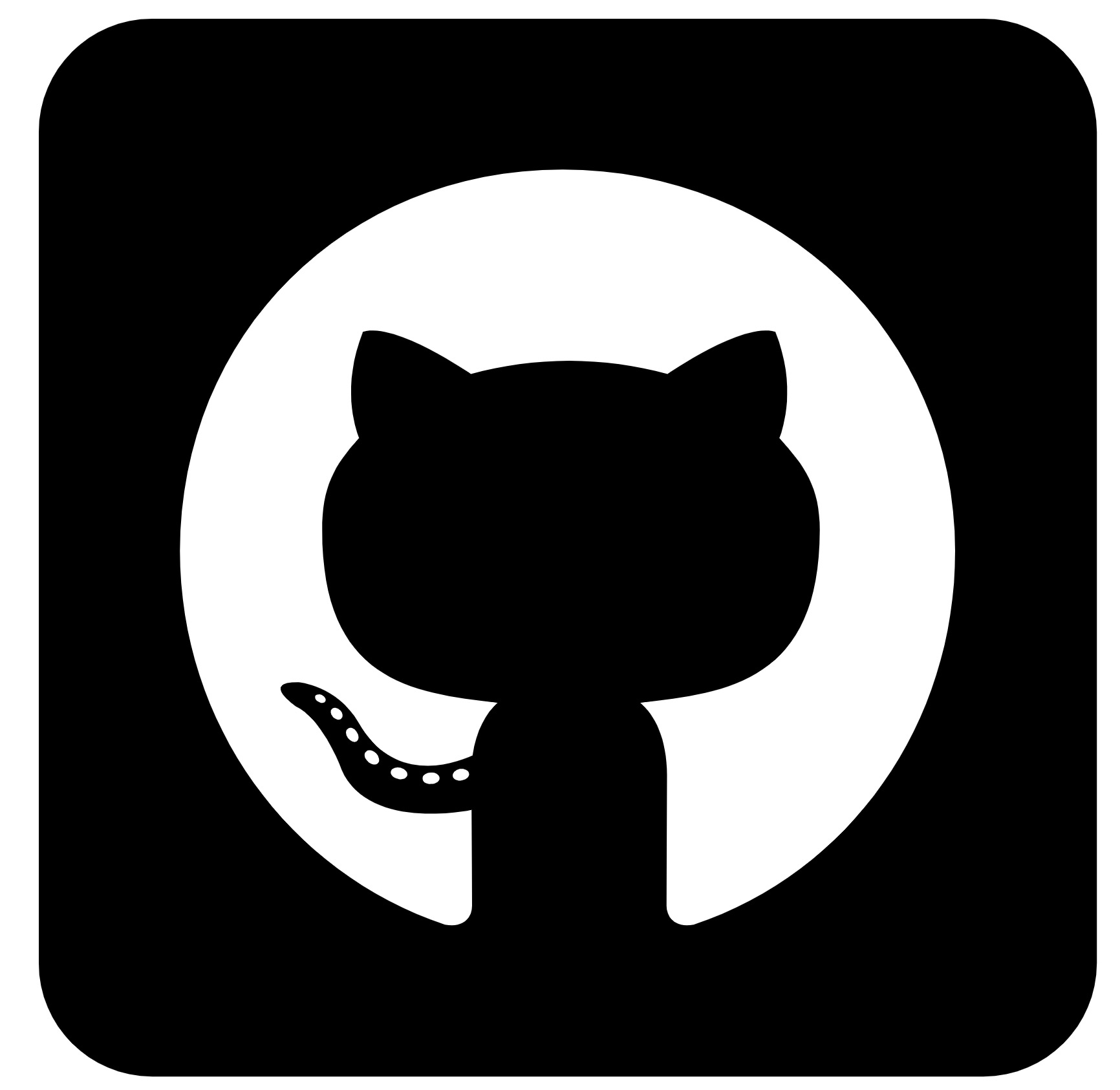}}\href{https://kaka0910.github.io/}{\includegraphics[height=7pt]{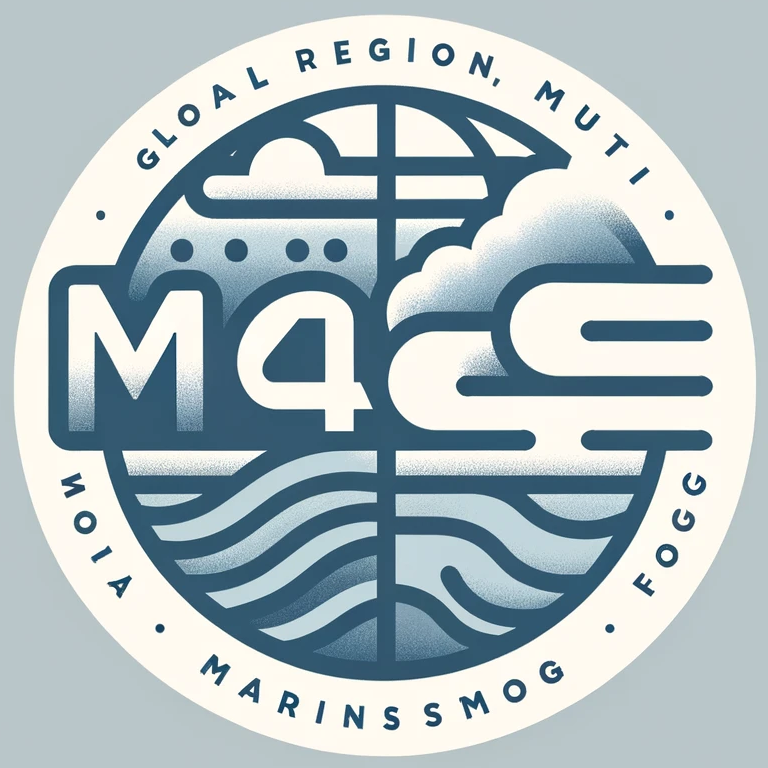}} to develop ML-driven marine fog solutions and mitigate adverse impacts on human activities.

\end{abstract}

\section{Introduction}
\label{sec:introduction}

Marine fog is a complex and hazardous meteorological phenomenon that occurs in the lower atmosphere over oceanic regions, primarily driven by temperature inversion and water vapor condensation~\cite{seafog_review_1,seafog_review_2}. During marine fog events, numerous water droplets and ice crystals are suspended in the atmospheric boundary layer, reducing horizontal visibility to less than one kilometer. This considerably disrupts navigation, port activities, and offshore operations~\cite{meteo_seafog_1,cctv_seafog}, particularly in economically prosperous coastal cities. However, unlike hurricanes and tsunamis~\cite{hurricanes_1,hurricanes_2} which have long garnered attention, research on the detection and forecasting of marine fog is still in its early stages and faces many challenges such as variable shapes and indistinct boundaries.

Over recent years, these challenges have spurred intense activity from both the meteorological and Machine Learning~(ML) communities. On one hand, traditional meteorological researchers monitor marine fog by analyzing brightness temperature differences across detection bands~\cite{seafog_traditional_1, seafog_traditional_2} or through physical mechanism analysis~\cite{seafog_physics}. On the other hand, ML offers different ways to calculate and interpret meteorological data. Both the timeliness and accuracy of marine fog detection and forecasting have improved through well-designed data-driven neural network models~\cite{seafog_dlinknet, seafog_scselinknet, eca_transunet, dual_branch}. 

Although these ML methods have demonstrated their potential, developing and benchmarking a practical marine fog model remains challenging for researchers or laboratories without a meteorological background and resources due to the lack of a curated, specialized dataset in the field. The few truly publicly accessible marine fog datasets~\cite{my_bmvc, dual_branch} have several inherent limitations. Firstly, the value of diversified data has not been adequately considered. Due to differences and delays in the data assimilation process, they have abandoned the utilization of multiple data sources and only provide single satellite remote sensing data. Secondly, these datasets fail to cover the full range of marine fog event processes, which is crucial for understanding the mechanisms of marine fog formation and dissipation. Finally, these datasets are always based on a single marine region, limiting the applicability of models and the development of foundational marine fog models. A dataset that avoids the aforementioned limitations is crucial for the advancement of the field.

\begin{figure}[t] 
\centering 
\includegraphics[width=\textwidth]{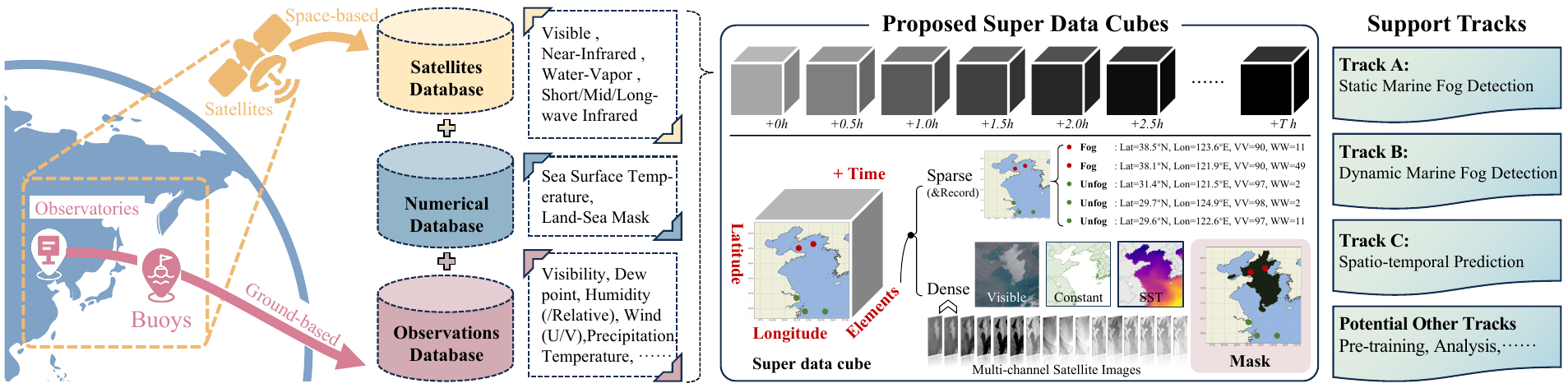} 
\caption{The overall construction flowchart of the proposed super data cubes in M4Fog is as follows. A variety of meteorological data related to marine fog detection and forecasting is obtained from geostationary satellites, numerical analysis, and observation databases. The super data cube is then constructed along four dimensions: elements, latitude, longitude, and time. According to visibility (VV) and present weather (WW) from observatories, the presence and absence of marine fog are determined.  Subsequently, M4Fog can support multiple meaningful tracks, including static/dynamic marine fog detection and spatio-temporal forecasting.} 
\label{Fig_construction} 
\end{figure}

In this paper, we introduce \textbf{M4Fog}, a global \textbf{M}ulti-regional, \textbf{M}ulti-modal, and \textbf{M}ulti-stage \textbf{M}arine Fog dataset, which is currently the most comprehensive dataset for marine fog detection and forecasting, aimed at establishing connections between the ocean and atmosphere. It encompasses thousands of marine fog events from 2015 to 2024, covering 15 major fog-prone regions worldwide. As shown in Fig.~\ref{Fig_construction}, M4Fog includes multi-modal data related to marine fog: remote sensing data from four series of geostationary satellites; reanalysis data of sea surface temperatures~\cite{sst}, which are commonly considered to be associated with the formation of marine fog; and visibility data from ground-based observation stations. To facilitate its application, we propose the concept of "super data cube", encompassing the four dimensions of elements, latitude, longitude, and time. M4Fog has approximately 68,000 "super data cubes", each with a temporal resolution of half an hour, a spatial resolution of 1 kilometer, and a size of $1024 \times 1024$. Among them, over 11,600 typical cubes are meticulously annotated at the pixel level by meteorological experts to specify the location and shape of marine fog. Based on the M4Fog dataset, we have defined and explored three tracks: static marine fog detection (Track A), dynamic marine fog detection (Track B), and spatio-temporal prediction of remote sensing cloud images (Track C). A series of experiments are conducted to benchmark a wide range of ML methods~\cite{unet, vit, unetformer, my_rs, predrnn, openstl} and demonstrate the value of the three "Multi" characteristics. The datasets and implementations are accessible via GitHub  \href{https://github.com/kaka0910/M4Fog}{\includegraphics[height=8pt]{Figs/github.jpg}} or homepage \href{https://kaka0910.github.io/}{\includegraphics[height=8pt]{Figs/M4Fog.png}}.

By building M4Fog, a new generation of specialized and user-friendly marine fog datasets, our goal is to attract more ML researchers to this field and promote the exploration of foundational patterns for marine fog detection and forecasting. Ultimately, this will significantly enhance the quality and reliability of meteorological services for marine operations.

\section{Related work}

\begin{table}[t]
\centering
\caption{The comparisons with other marine fog benchmarks or datasets, focusing on tracks (i.e., numbers, w/o pixel-level), regionality (i.e., specific vs. multi-regions, abbreviated as (Sea area, Continent) referring to Appendix~\ref{sec:data_details} ), temporality (e.g., seasonal, annual), variety of training/test (i.e.,satellites vs. observations vs. others), continuity and w/o open.}
\label{tab:datasets_review}
\resizebox{\textwidth}{!}{%
\begin{tabular}{l|cc|cc|cc|ccc|c|c}
    \toprule
    \multirow{2}{*}{\textbf{Datasets}}  & \multicolumn{2}{c|}{\textbf{Tracks}} & \multicolumn{2}{c|}{\textbf{Regionality}} & \multicolumn{2}{c|}{\textbf{Temporality}} & \multicolumn{3}{c|}{\textbf{Variety} \small{(training/test)}}  & \textbf{Conti-} & \multirow{2}{*}{\textbf{Open}} \\
     & \multicolumn{2}{c|}{\tiny{(Num./Pixel-level)}} & \small{Specific} & \small{Multi} & \small{Seasonal} & \small{Annual} & \small{Sats.} & \small{Obs.} & \small{Others} & \textbf{nuity} &  \\
     
    \midrule
    \textbf{Nilo et al.} (2018)~\cite{meteo_seafog_2} & 1 & \ding{51} &  \ding{51} \tiny{(PA,EU)} &  &  & \ding{51} & \ding{51} / \ding{51} & - / \ding{51} & - / - &  &  \\
    \textbf{Huang et al.} (2021)~\cite{my_rs} & 1 & \ding{51} &  \ding{51} \tiny{(YB,AS)} &  & \ding{51} & \ding{51} & \ding{51} / \ding{51} & - / \ding{51} & - / - &  &  \\
    \textbf{Mahdavi et al.} (2021)~\cite{goes_seafog_1} & 1 & \ding{51} &  \ding{51} \tiny{(GB,NA)} &  &  &  & \ding{51} / \ding{51} & - / \ding{51} & \ding{51} / - &  &  \\
    \textbf{Zhou et al.} (2022)~\cite{dual_branch} & 1 & \ding{51} & \ding{51} \tiny{(YB,AS)} &   & \ding{51} & \ding{51} & \ding{51} / \ding{51} & - / - & - / \ding{51} & \ding{51} &  \ding{51} \\
    \textbf{Tao et al.} (2022)~\cite{seafog_self} & 1 &  & \ding{51} \tiny{(YB,AS)} &   &  & \ding{51} & \ding{51} / \ding{51} & - / - & - /  &  &  \\ 
    \textbf{Su et al.} (2023)~\cite{my_bmvc} & 2 & \ding{51} & \ding{51} \tiny{(YB,AS)} &   & \ding{51} & \ding{51} & \ding{51} / \ding{51} & - / - & - / - &  & \ding{51} \\ 
    \textbf{Bari et al.} (2023)~\cite{meteo_seafog_1} & 2 &  & \ding{51} \tiny{(MA,AF)} &   & \ding{51} & \ding{51} & \ding{51} / \ding{51} & - / - & - / - & \ding{51} &  \\
    
    \midrule
    \textbf{Ours}  & 3 & \ding{51} & \ding{51}  & \ding{51} & \ding{51} & \ding{51}  & \ding{51} / \ding{51} & \ding{51} / \ding{51} & \ding{51} / \ding{51} & \ding{51} & \ding{51} \\
    \bottomrule
\end{tabular}%
}
\end{table}

With the diversification of meteorological data collection and the advancement of machine learning data digitization, multiple sea fog detection and forecasting datasets~\cite{meteo_seafog_2,my_rs,goes_seafog_1,dual_branch,seafog_self,my_bmvc,meteo_seafog_1} have been proposed in the field to indicate the presence or absence of marine fog. These datasets have promoted early research, and even some ML-based sea fog detection and forecasting methods~\cite{seafog_dlinknet, dual_branch, eca_transunet} have surpassed traditional meteorological methods~\cite{seafog_traditional_1,seafog_traditional_2}.

However, existing datasets often resemble a "toy" imitation up to this point. They are confined to simplistic scenarios to meet the research goals of ML, without being task-oriented or fully exploiting the characteristics of marine fog and available data. Specifically:
(1) \textbf{Variety deficiency}. Most current methods rely on satellite data corresponding to one maritime area, such as the Himawari series~\cite{seafog_dlinknet, my_grsl} or FengYun series satellites~\cite{FY4A_seafog_1,fy3d_seafog_1} for studies in the Yellow and Bohai Sea. Although some studies~\cite{meteo_seafog_2,seafog_dlinknet, my_cja} have proposed using observation stations or active satellites for validation, this is also not allowed to comprehensively simulate marine fog through diverse data during training process. Therefore, the first key breakthrough of M4Fog lies in incorporating diverse meteorological data related to marine fog.
(2) \textbf{Temporal discreteness}. It is our contention that the discontinuous samples generated by most data construction methods~\cite{meteo_seafog_2, seafog_self,goes_seafog_1} are inconsistent with the persistence of marine fog. Although MoANet~\cite{my_jstars} proposes the positive impact of temporal motion vector, the construction of data sequences remain relatively rough, lacking a favorable analytical basis. Therefore, M4Fog not only considers continuity as one of the principles for dataset constructions but also designs a dynamic marine fog detection track to further explore the effectiveness of temporal features.
(3) \textbf{Regional specificity}. It's not hard to notice that most methods remain limited to their respective areas of focus. Even contemporaneous Zhou et al.~\cite{dual_branch} concentrate on marine fog in the Yellow and Bohai Sea in Asia, while Mahdavi et al.~\cite{goes_seafog_1} focus on areas near the Great Banks in North America. This lacks discussions on the influence of global-scale phenomena like sea surface temperature, or even preparations for future global warming challenges. In contrast, M4Fog extends the regional restrictions of previous datasets by collecting and integrating data from globally frequent fog occurrence regions.
(4) \textbf{Track monotony}. Most existing datasets are mostly limited to a single task agent, predominantly focused on marine segmentation~\cite{seafog_dlinknet,dual_branch,goes_seafog_1,meteo_seafog_2}, thereby constraining the expansion of boundaries, such as pre-training~\cite{my_bmvc} and weak-supervised learning~\cite{my_tgarss,my_vcip} in fog-related tasks. This hinders the emergence of fresh forces in both the meteorological and ML communities. Therefore, three meaningful ML tracks is defined paving the way for potential avenues.
In addition, we present a detailed comparison of benchmarks or datasets prone to marine fog detection and forecasting as shown in Table.~\ref{tab:datasets_review}. It is worth noting that most current research is not open-source thereby limiting the exchange of technology in the meteorology and technology communities.

\begin{figure}[t]
  \begin{minipage}{0.55\textwidth}
    \includegraphics[width=\linewidth]{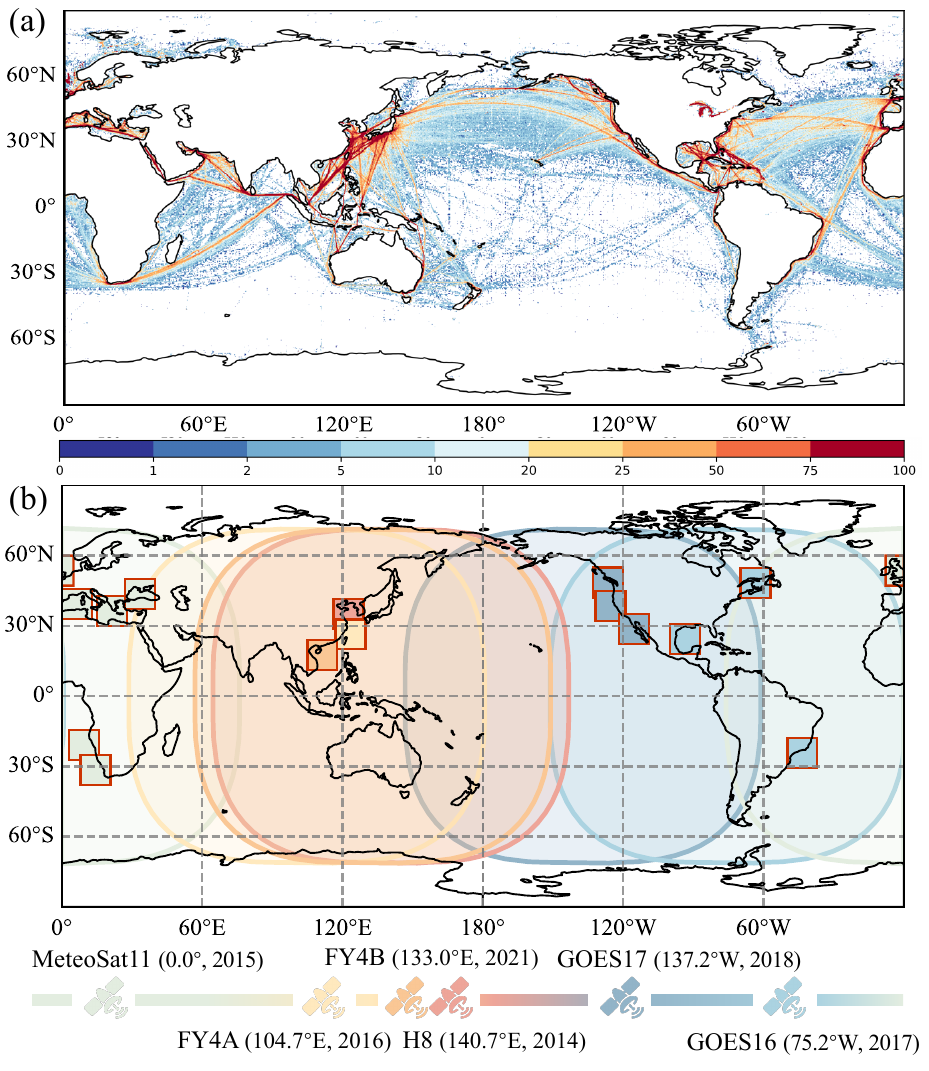} 
    \caption{(a) Analysis of global marine fog occurrence frequency based on ICOADS records during 2015-2024. (b) Schematic diagram of part of the global meteorological satellite detection system and selected areas.}
    \label{fig:merge}
  \end{minipage}
  \hspace{0.02\textwidth}
  \begin{minipage}{0.40\textwidth}
  \small{
  \adjustbox{valign=t}{%
      \begin{tabularx}{\textwidth}{X|l}
        \toprule
        Meteorological Data & Num. \\ 
        \midrule
        \tiny{\textbf{FengYun-4 Series}} &  \\
        \tiny{Yellow Sea/ Bohai Sea/ East China Sea/ North China Sea and adjacent marine regions} &  $\approx 1.8k $ \\
        \tiny{Unlabeled Data cubes} & $\approx 21k $ \\
        \tiny{\textbf{GOES Series}} &  \\
        \tiny{Labeled Victoria/ San Francisco/ Brazil / Baja California/ Gulf of St.Lawrence/ Gulf of Mexico and adjacent maritime regions} &  $\approx 3.1k $ \\
        \tiny{Unlabeled Data cubes} & $\approx 12k $ \\
        \tiny{\textbf{Himawari-8/9 Series}} &  \\
        \tiny{Yellow Sea/ Bohai Sea/ Sea of Japan/ North Korea and South Korea and adjacent maritime regions} &  $\approx 2.5k $ \\
        \tiny{Unlabeled Data cubes} & $\approx 20k $ \\
        \tiny{\textbf{MeteoSat-2rd Series}} &  \\
        \tiny{Mediterranean Sea/ South Africa/ English Channel and adjacent maritime regions} &  $\approx 4.2k $ \\
        \tiny{Unlabeled Data cubes} & $\approx 15k $ \\
        \midrule
        \multicolumn{2}{l}{Satellites: \tiny{Labeled $\approx 11.6k$ / Unlabeled $\approx 68k$} }  \\
        \midrule
        ICOADS \tiny{(2015-2024)} & $\approx 95m $ \\
        \midrule
        Observed records  & $\approx 14k $ \\
        \midrule
        Global SST \tiny{(2018-2024)} & $\approx 2.2k $ \\
        \bottomrule
      \end{tabularx}
    }
    \captionof{table}{Various meteorological data used with regional information, as well as the volume of labeled and unlabeled data units.}
    \label{tab:volume}
  }
  \end{minipage}
\end{figure}

\section{The M4Fog dataset}

\textbf{Overview Information.} M4Fog is a global \textbf{M}ulti-regional, \textbf{M}ulti-modal, and \textbf{M}ulti-stage \textbf{M}arine fog detection and forecasting dataset sourced from a variety of meteorological data including geostationary meteorological satellites, numerical analysis, and observation stations. One of the advantages of M4Fog is its extensive temporal and spatial coverage. It spans nearly a decade, from 2015 to 2024, and covers 15 coastal regions worldwide with high marine fog frequency and significant human activity. All dense and sparse data have been processed into regular "super data cubes" to facilitate ML model training. The dataset contains totaling up to 68,000 cubes, of which 11,600 have pixel-level annotations for marine fog provided by meteorological experts. More detailed quantity information about the data involved in our M4Fog can be found in Table.~\ref{tab:volume}.

\textbf{Construction Principles.} The construction of M4Fog is based on two fundamental objectives. (1) Research Objective: the quality, quantity, and diversity of the data should meet the requirements of machine learning. (2) Practical Objective: the dataset composition should reflect real-world marine fog meteorological service scenarios to accelerate the deployment of technology. To achieve these two objectives, we have established six construction principles: \textbf{Variety}, \textbf{Continuity}, \textbf{Regionality}, \textbf{Temporality}, \textbf{Usability}, and \textbf{Veracity}. They are ultimately reflected in M4Fog as the four "M"s, specifically as follows:

\textbf{\textcolor{brown}{M}ulti-modal.} There are three modalities of data in M4Fog: satellite remote sensing data, observation station data, and numerical analysis data. Firstly, the satellite remote sensing data, referred to as the "sky eye"~\cite{data_review_1,data_review_2}, meets the spatial and temporal resolution requirements. These data come from the Global Meteorological Satellite Detection System, specifically including the Fengyun-4 series satellites, Himawari-8/9 series satellites, GOES series satellites, and Meteosat series satellites, as shown in Fig.~\ref{fig:merge}~(b) in different colors. Secondly, the marine surface-level observation station data, acting as a "frontline reporter", supplements with detailed meteorological information related to marine fog. Since most observation stations do not directly observe marine fog phenomena, we used expert knowledge for deducing meteorological elements to marine fog phenomena. For instance, according to official documentation and related studies on ICOADS data~\cite{icoads}, the primary meteorological elements or detection values/phenomena related to marine fog are visibility (VV) and the present weather (WW). The criteria for fog determination are $VV \leq 94, WW<60$, mainly to exclude the impact of precipitation and other weather conditions on visibility and fog. The criteria for identifying marine fog vary among different observation stations, and specific identification experiences are recorded in Appendix~\ref{sec:data_details}. Lastly, the numerical analysis data, including geographical constant data and sea surface temperature data, further enrich data features from different scales and perspectives. All data are aligned by timestamp and location. 

\textbf{\textcolor{brown}{M}ulti-stage.} The multi-stage aspect of M4Fog is reflected on two temporal scales: the lifecycle of a marine fog and the annual cycle. Marine fog is a continuously changing weather phenomenon that includes multiple stages such as formation, expansion, advection, maintenance, and dissipation. We argue that the temporal information contained in continuous data is of significant value for the detection and forecasting of marine fog. Therefore, when collecting data, we ensured the completeness of each marine fog event and the continuity of time as much as possible to allow for the study of temporal information. Considering the varying temporal resolutions of different satellites, we standardized our data collection to a half-hour interval. Additionally, according to the analysis supplied in the Appendix~\ref{sec:analysis_season}, the occurrence rate of marine fog varies by quarter and season. For example, marine fog in the Yellow Sea/Bohai Sea occurs mostly in spring and summer, while in the South China Sea, it occurs mainly in winter and spring. Therefore, M4Fog also provides data from multiple stages throughout the year for detailed study.

\textbf{\textcolor{brown}{M}ulti-regional.} Marine fog is a global phenomenon occurring from coastal areas~\cite{goes_seafog_1, fy3d_seafog_1, cctv_seafog, meteo_seafog_1} to polar regions~\cite{arctic_fog, arctic_fog_1}, excluding tropical oceans, provided the necessary temperature and humidity conditions are met. Lower sea temperatures causing air to cool will lead to the marine fog, suggesting that the occurrence of marine fog is greatly influenced by regions. To better handle global marine fog, we extract almost 9.5 million records from all ICOADS~\cite{icoads} data spanning 2015-2024, which clearly observe or infer the presence or absence of marine fog globally. These records are accumulated into a global grid of 0.25° $\times$ 0.25° for frequency tallying, followed by statistical analysis and visualization, as illustrated in Fig.~\ref{fig:merge} (a). It is evident that human maritime activities are mostly concentrated near coasts or along specific shipping routes, as observations of marine fog are distributed accordingly. Based on this, We select 15 regions around the world where marine fog frequently occurs and is highly correlated with human activities, represented by differently colored blocks in Fig.~\ref{fig:merge} (b). We hope that our multi-regional M4Fog will start the study of global generalization and the development of foundational marine fog models.

\textbf{\textcolor{brown}{M}arine Fog Super Data Cube.} To enhance the usability of M4Fog, we propose the concept of the "super data cube", which encompasses the four dimensions of element, latitude, longitude, and time. We have reconstructed all the collected data into 68,000 marine fog super data cubes as illustrated in the Fig.~\ref{Fig_construction}. Specifically, the dense satellite remote sensing data, sea surface temperature data, and constant data are first concatenated by element dimension. Then, the sparse observation station data is overlaid onto the cube by latitude and longitude. Besides, we select over 11,600 typical data cubes and provide true-color and natural-color images to meteorological experts for pixel-level annotation. The marine fog masks are also concatenated into these cubes along the element dimension. The methods for synthesizing visualized images for different satellites are detailed in the Appendix~\ref{sec:data_details}.

\section{Machine Learning Tracks}

\subsection{Track A: Static Marine Fog Detection}
\label{tracka}

\textbf{Track Definition.} Static marine fog detection refers to identifying and delineating marine fog areas at a specific moment based on super data cubes from that moment. It is a fundamental application of machine learning in this field. Specifically, given a timestamp and marine region, the machine learning model is trained to map the data $ \mathcal{X} \in \mathbb{R}^{C \times H \times W}$ to a binary mask $ \mathcal{M} \in \mathbb{R}^{ H \times W}$, where each pixel value $ \mathcal{M}_{i,j} $ indicates the presence or absence of marine fog at pixel location $(i,j)$.


\begin{table}[t]
\centering
\caption{The quantitative results for Track A Static Marine Fog Detection experiments using proposed sub-dataset collected from Yellow and Bohai Sea using Fengyun-4A and Himawari-8/9 satellites.}
\label{tab:Track_static}
\resizebox{\textwidth}{!}{%
\begin{tabular}{c|ccc|cc|ccc|cc}
    \toprule
     & \multicolumn{5}{c|}{\textbf{FY-4A}} & \multicolumn{5}{c}{\textbf{H-8/9}} \\
    \cmidrule{2-6}
    \cmidrule{7-11}
    \textbf{Method} & \multicolumn{3}{c|}{Dense-metrics} & \multicolumn{2}{c|}{Sparse-metrics} & \multicolumn{3}{c|}{Dense-metrics} & \multicolumn{2}{c}{Sparse-metrics} \\
    & CSI $\uparrow$ & mIoU  $\uparrow$ &  Acc $\uparrow$ & TS $\uparrow$ & F1-score $\uparrow$ &  CSI $\uparrow$ & mIoU $\uparrow$ &  Acc $\uparrow$ &  TS $\uparrow$ & F1-score $\uparrow$ \\
    \midrule
    Deeplabv3p \cite{deeplabv3+} & 44.07 & 70.48 & 66.88 & 0.3650 & 0.5348 & 51.83 & 74.73 & 68.61 & 0.3866 & 0.5577 \\
    UNet \cite{unet}             & 45.77 & 71.29 & 73.42 & 0.3871 & 0.5581 & 54.48 & 76.15 & 70.60 & 0.4433 & 0.6143 \\
    UNet++ \cite{unetpp}         & 45.79 & 71.30 & 73.38 & 0.4254 & 0.5969 & 55.71 & 76.78 & \textbf{73.32} & 0.4383 & 0.6095\\
    ABCNet \cite{abcnet}         & 46.46 & 71.68 & 73.44 & 0.3927 & 0.5640 & \underline{57.36} & \underline{77.71} & 70.20 & 0.4633 & 0.6332 \\
    BANet \cite{banet}           & 46.67 & 68.72 & 72.11 & 0.3638 & 0.5335 & 55.11 & 76.54 & 67.55 & 0.4408 & 0.6119 \\
    Unetformer \cite{unetformer} & 48.66 & 72.93 & 78.03 & 0.4166 & 0.5881 & 54.63 & 76.26 & 68.91 & 0.4465 & 0.6173 \\
    ViT-Base \cite{vit}          & \underline{52.77} & \underline{75.08} & \textbf{79.38} & \underline{0.5263} & \underline{0.6897} & 57.03 & 77.50 & \underline{72.51} & \underline{0.4912} & \underline{0.6588} \\
    Dlink-ViT-Base \cite{my_rs} & \textbf{56.93} & \textbf{77.38} & \underline{78.30} & \textbf{0.5494} & \textbf{0.7092} & \textbf{60.25} & \textbf{79.24} & 71.92 & \textbf{0.5903} & \textbf{0.7423} \\    
    \bottomrule
\end{tabular}%
}
\end{table}

\begin{figure}[t] 
\centering 
\includegraphics[width=\textwidth]{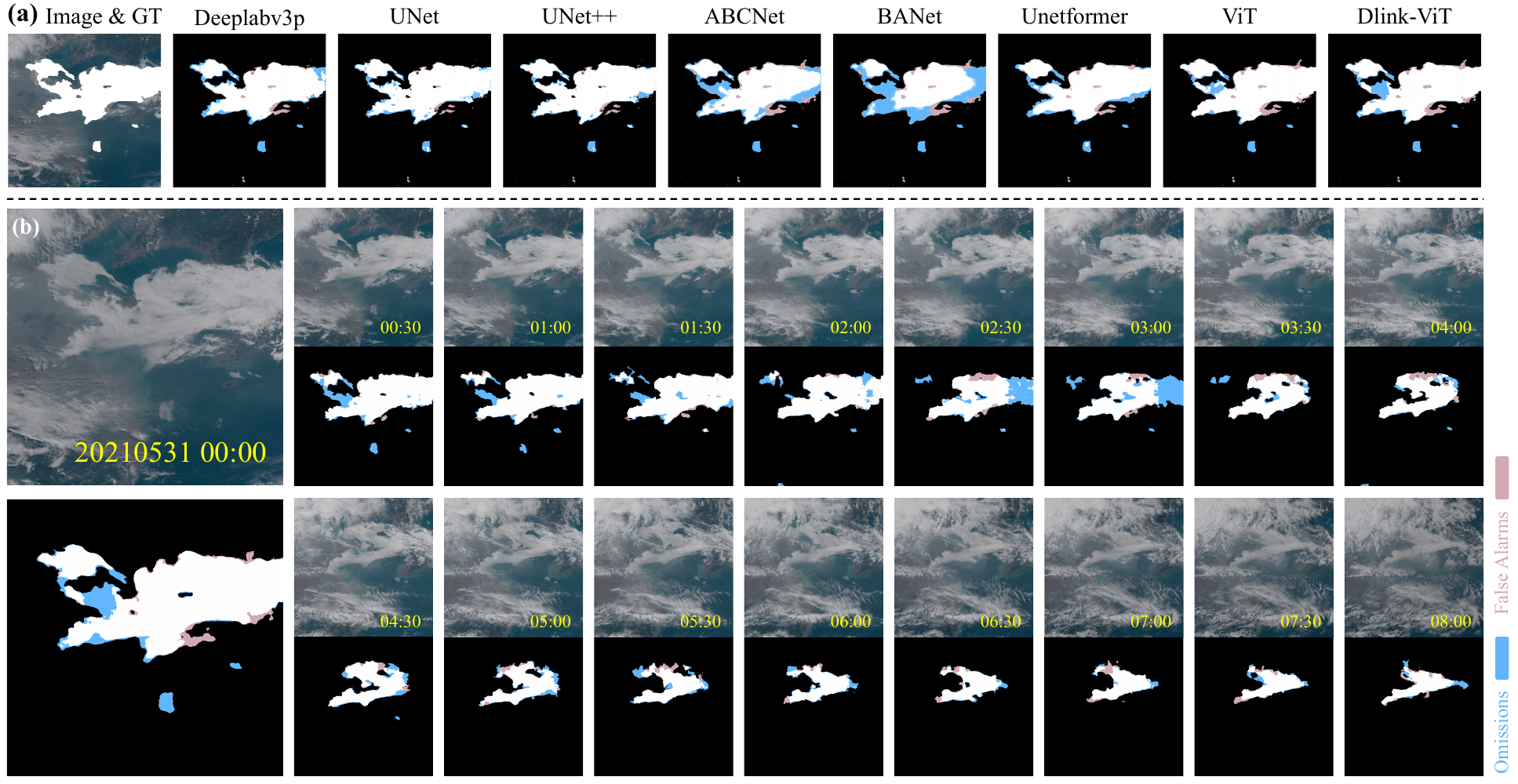} 
\caption{The qualitative visualization on Track A Static Marine Fog Detection includes (a) a comparison between different methods, and (b) an illustrative instance of uninterrupted marine fog surveillance on March 24, 2021 in Yellow and Bohai Sea based on H8/9 satellite images. } 
\label{Fig_tracka} 
\end{figure}

\textbf{Track and Experiment Setup.} Considering that different satellites have varying numbers of detection channels, we divided the annotated super data cubes into four independent sub-datasets based on satellite type, and further split them into training and test sets by year. To comprehensively evaluate the models, we employed five different evaluation metrics: the Critical Success Index (CSI), mean Intersection over Union (mIoU), and Accuracy (Acc)~\cite{csi} commonly used in deep learning, as well as the Threat Score (TS) and F1 score based on sparse ground-based observations in meteorology. Due to space limitations, we only report the situation and experimental results for the Fengyun-4A and Himawari-8/9 sub-datasets in the main text. The Fengyun-4A sub-dataset and the Himawari-8/9 sub-dataset contain 1,724 and 1,802 data, respectively, with data before 2021 used as the training set and data after 2021 used as the test set. Details of other sub-datasets and the formulas for the evaluation metrics can be found in the Appendix~\ref{sec:tracka_details}.

In the experiments, eight classic vision-based models are evaluated, including CNN-based models~\cite{deeplabv3+, unet, unetpp, abcnet} and ViT-based models~\cite{banet, unetformer, my_rs, vit}. Additionally, we conduct further experiments on the multi-modal data within the super data cubes to explore the effectiveness of sea surface temperature data and constant data in the static marine fog detection track.

\textbf{Results and Analysis.} Table.~\ref{tab:Track_static} and Fig.~\ref{Fig_tracka} show the quantitative and qualitative results of all vision-based neural network models, respectively. We find that the ViT-based models generally outperform the CNN-based models, with Dlink-ViT-Base~\cite{my_rs} significantly outperforms other models on both satellites. On the other hand, Table.~\ref{tab:super_data_cube} shows the results when concatenating sea surface temperature and constant data with satellite data. It can be seen that even simple dimensional concatenation can lead to significant performance improvements, exhibiting the same effect across different models. This result effectively underscores the value of multi-modal data in our super data cube for static marine fog detection.


\begin{figure}[t]
  \begin{minipage}{0.45\textwidth}
    \captionof{table}{The effectiveness of proposed super data cube constructions on H8/9 sub-dataset.}
    \begin{adjustbox}{valign=t}
        \resizebox{\textwidth}{!}{%
          \begin{tabular}{c|c|cc}
          \toprule
          \textbf{Method} & \textbf{Settings} & \textbf{CSI} $\uparrow$ & \textbf{Acc} $\uparrow$ \\
          
          \midrule
          UNet \cite{unet}             & +Cons. & 55.99 \tiny{\textcolor{red}{+1.51}} & 70.74 \tiny{\textcolor{red}{+0.14}} \\
          ViT-Base \cite{vit}          & +Cons. & 58.87 \tiny{\textcolor{red}{+1.94}} & 73.85 \tiny{\textcolor{red}{+1.34}} \\
          
          \midrule
          UNet \cite{unet}             & +SST  & 56.46 \tiny{\textcolor{red}{+1.98}} & 56.46 \tiny{\textcolor{red}{+1.98}} \\
          ViT-Base \cite{vit}          & +SST  & 59.00 \tiny{\textcolor{red}{+1.97}} & 74.56 \tiny{\textcolor{red}{+2.05}} \\
          
          \midrule
          UNet \cite{unet}             & +Cons.+SST & 55.89 \tiny{\textcolor{red}{+1.41}} & 72.28 \tiny{\textcolor{red}{+1.68}} \\
          ViT-Base \cite{vit}          & +Cons.+SST &59.26 \tiny{\textcolor{red}{+2.23}} & 75.62 \tiny{\textcolor{red}{+3.11}} \\
          \bottomrule
          \end{tabular}
        }
    \end{adjustbox}
    \label{tab:super_data_cube}
  \end{minipage}
  \hspace{0.01\textwidth}
  \begin{minipage}{0.50\textwidth}
    \captionof{table}{The quantitative results for Dynamic marine fog detection on H8/9 sub-dataset when N=1.}
    \begin{adjustbox}{valign=t}
        \resizebox{\textwidth}{!}{%
          \begin{tabular}{c|c|cc}
          \toprule
          \textbf{Method} & \textbf{Settings} & \textbf{CSI} $\uparrow$ & \textbf{Acc} $\uparrow$ \\
          \midrule
          UNet~\cite{unet}              & -     & 54.17 & 69.37 \\
          Temporal UNet~\cite{temporal_unet} & -     & 53.86 \tiny{\textcolor{green}{-1.31}} & 71.70 \tiny{\textcolor{red}{+2.33}} \\
          UNet~\cite{unet}               & Dual-ways  & 55.75 \tiny{\textcolor{red}{+1.58}} & 71.02 \tiny{\textcolor{red}{+1.65}} \\
          MoANet~\cite{my_jstars} & +Motion.  & 54.58 \tiny{\textcolor{red}{+0.41}} & 68.21 \tiny{\textcolor{green}{-1.16}} \\
          
          \midrule
          Dlink-ViT-Base~\cite{my_rs} & -        & 60.48 & 71.91 \\
          Dlink-ViT-Base~\cite{my_rs} & Dual-ways& 61.90 \tiny{\textcolor{red}{+1.42}} & 75.81 \tiny{\textcolor{red}{+3.90}} \\
          \bottomrule
          \end{tabular}
        }
    \end{adjustbox}
    \label{tab:example_table1}
  \end{minipage}
\end{figure}

\subsection{Track B: Dynamic Marine Fog Detection}
\label{trackb}

\textbf{Track Definition.} In contrast to static marine fog detection, dynamic marine fog detection refers to identifying and delineating marine fog areas at a specific moment by utilizing continuous super data cubes that incorporate data from a preceding time period, given that the persistence of multi-stage marine fog events is a key characteristic for identifying, particularly in situations where status cloud and marine fog are confused. Given a timestamp, a marine region, and $N$ previous historical data moments spaced by a time step $t_{step}$, one super data cube $ \mathcal{X} \in \mathbb{R}^{(N+1) \times C \times H \times W}$ with a machine learning model is constructed to predict the current marine fog binary mask $M \in \mathbb{R}^{ H \times W}$. 


\textbf{Track and Experiment Setup.} Building upon the sub-dataset construction of Track A, for each sample, we incorporate the super data cubes from the previous $N$ moments leading up to the current timestamp to form a new instance. Finally, four independent sub-datasets are created based on satellite types and the value of $N$. There are two evaluation metrics are chosen in this track: the Critical Success Index (CSI) and Accuracy (Acc)~\cite{csi}. We only present experimental results based on the Himawari-8/9 satellites sub-dataset with 1,696 super data cubes when $N=1$ in the main text.

In the experiments, Temporal UNet and Dual-ways structured models~\cite{temporal_unet} are introduced into this track, with UNet~\cite{unet} and DlinkViT-Base~\cite{my_rs} serving as the foundational backbones comparing the quantitative results of eight baselines in Track A. We subsequently follow MoANet~\cite{my_jstars} to utilize intermediate variables (such as motion or optical flow) derived from continuous data sequences to further validate the effectiveness of temporal features. Experimental results for other sub-datasets and detailed experimental procedures are provided in the Appendix~\ref{sec:trackb_details}.


\textbf{Results and Analysis.} The quantitative results of the dynamic marine fog detection experiments are shown in Table.~\ref{tab:example_table1}. Specifically, Line 1 and Line 5 present the evaluation results for UNet~\cite{unet} and Dlink-ViT~\cite{my_rs} models trained on Track A but validated on Track B's sub-dataset, respectively. In addition, simply expanding the model input dimensions by concatenating temporal data yields minimal gains, indicating the necessity of designing separate pathways for different times and types of data or features according to the Dual-ways structured experimental results. Overall, we can draw a conclusion that it is effective to incorporate past continuous data cubes and utilizing intermediate temporal variables as performance improvements are observed across different backbones.

\subsection{Track C: Spatio-Temporal Prediction for Cloud Images}
\label{trackc}

\textbf{Track Definition.} The continuous multi-stage super data cubes not only provide temporal information for marine fog detection but also allow us to perform spatio-temporal tasks. Specifically, considering the importance of prediction, we have defined a more challenging and valuable track for M4Fog: spatio-temporal prediction for cloud image forecasting. Given a continuous sequence of data cube $ \mathcal{X}^{t,T}=\{x^i\}^t_{t-T+1} $ that traces back from time $t$ to past time $T$, the goal is to predict the upcoming sequence of data cube $ \mathcal{Y}^{t+1,t+T'}=\{x^i\}^{t+T'}_{t+1} $ until time $T'$. Each data cube $x^i \in \mathbb{R}^{C \times H \times W}$ of sequence comprises $C$ channels with height $H$ and width $W$. During the model learning process, the sequence data cubes are transformed into four-dimensional tensors $ \mathcal{X}^{t,T} \in \mathbb{R}^{T \times C \times H \times W} $ and $ \mathcal{Y}^{t+1, T'} \in \mathbb{R}^{T' \times C \times H \times W} $. Thus, the goal of spatio-temporal prediction model is to map $ \mathcal{X}^{t,T} \rightarrow \mathcal{Y}^{t+1, T'} $ by minimizing the discrepancy between the predictions and the ground truths. $C$, $T$ and $T'$ can be adaptively set according to the requirements of actual forecasting operations.


\begin{table}[t]
\centering
\caption{The performance on Track C Spatio-temporal prediction using continuous Himawari-8/9 satellite images ranging from single Yellow and Bohai Sea to combination with multi marine areas.}
\label{tab:ST_prediction}
\resizebox{\textwidth}{!}{%
\begin{tabular}{ll|lccccl|lccccl}
    \toprule
     \multicolumn{2}{l|}{\multirow{2}{*}{\textbf{Method}}}  & \multicolumn{6}{c|}{\textbf{Single-area}} & \multicolumn{6}{c}{\textbf{Multi-areas}} \\
    & & & MSE $\downarrow$ &  MAE $\downarrow$ & SSIM $\uparrow$ &  PSNR $\uparrow$ & & & MSE $\downarrow$ &  MAE $\downarrow$ & SSIM $\uparrow$ &  PSNR $\uparrow$ &\\
    \midrule
    ConvLSTM \cite{convlstm}  & & & 697.12 & 7540.91  & 0.7918  & 25.31 & & & 858.02 & 8458.56 & 0.7513 & 24.31 &\\
    PredRNN \cite{predrnn}    & & & \textbf{636.37} & \textbf{7067.29}  & \textbf{0.8090}  & \textbf{25.77} &&  & \textbf{800.69} & \textbf{7997.54} & \textbf{0.7702} & \textbf{24.68} &\\
    MIM \cite{mim}            & & & 668.42 & 7325.09  & \underline{0.8002}  & 25.52 & & & \underline{846.27} & 8387.89 & 0.7552 & \underline{24.38} &\\
    PhyDNet \cite{phydnet}    & & & 803.84 & 8447.91  & 0.7770  & 24.58 & & & 974.90 & 9264.14 & 0.7337 & 23.68 &\\
    \midrule
    SimVP-v2 \cite{simvpv2}   & & & 726.83 & 7644.08  & 0.7913  & 25.18 & & & 939.75 & 8756.57 & 0.7466 & 23.97 &\\
    Uniformer \cite{uniformer}& & & 679.37 & 7435.03  & 0.7988  & 25.42 & & & 871.40 & 8461.68 & \underline{0.7588} & 24.26 &\\
    VAN \cite{van}            & & & 728.04 & 7635.44  & 0.7908  & 25.18 & & & 932.17 & 8715.68 & 0.7458 & 24.00 &\\
    TAU \cite{tau}           & & & \underline{660.41} & \underline{7273.11}  & 0.7960  & \underline{25.54} & & & 855.29 & \underline{8339.58} & 0.7530 & 24.34 &\\
    \bottomrule
\end{tabular}%
}
\end{table}

\begin{figure}[t] 
\centering 
\includegraphics[width=\textwidth]{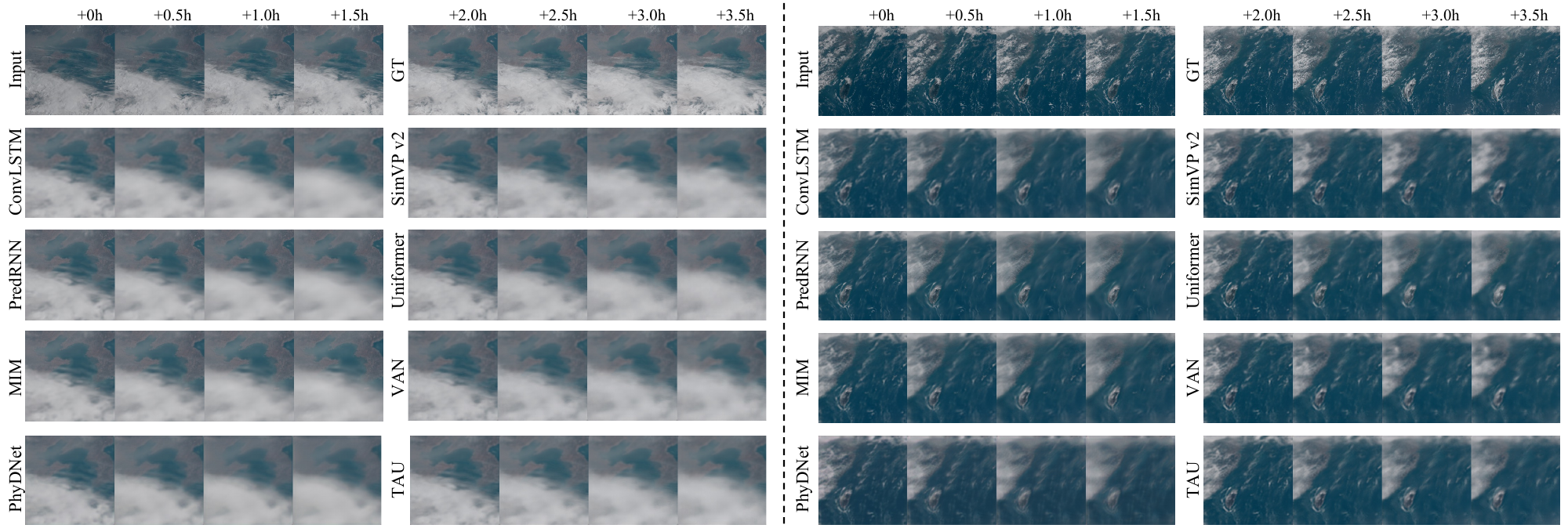} 
\caption{The qualitative visualization of Track C ranges from single Yellow and Bohai Sea (left) to a combination with multiple other marine areas (right) on continuous Himawari-8/9 satellite images.} 
\label{Fig_result_prediction} 
\end{figure}

\textbf{Track and Experiment Setup.} In this track, we construct eight sub-datasets using all the super data cubes, with each satellite having a single-region sub-dataset and a multi-region sub-dataset. The datasets are split by year to prevent test set leakage. We set $C=3$, and $T=T'=4$. To maximize data utilization, we merely allow overlapping data in the training set. As with traditional spatio-temporal prediction tasks, we also provide Mean Squared Error (MSE), Mean Absolute Error (MAE), Structural Similarity Index Measure (SSIM) and Peak Signal-to Noise Ratio (PSNR)~\cite{metrics} as evaluation metrics to measure model performance. Here, we use the Himawari-8/9 satellites as an example. We extract data from two regions, the Yellow and Bohai Sea and the East China Sea, obtaining 5,565 and 5,367 data cubes, respectively. For feasible visualization of the data and experimental results, we selected the visible bands (0.47 $\mu$m, 0.51 $\mu$m, 0.64 $\mu$m) from these data cubes. Data from the two regions are used to create 2,512 and 2,432 continuous sequences of length $4+4$, respectively. The 2021 year is also defined as a boundary, with data before 2021 serving as the training set and data after 2021 as the test set, resulting in 1,840 / 672 China Yellow-Bohai Sea and 1,760 / 672 East China Sea sequences, respectively.

For the experiments, the OpenSTL~\cite{openstl} framework is selected as the foundation. Following OpenSTL~\cite{openstl}, we categorize spatio-temporal prediction methods into two main types. One type are recurrent-based methods, including ConvLSTM~\cite{convlstm}, PredRNN~\cite{predrnn}, MIM~\cite{mim} and PhyDNet~\cite{phydnet}, etc. Another type are recurrent-free methods, including SimVP v2~\cite{simvpv2}, Uniformer~\cite{uniformer}, VAN~\cite{van} and TAU~\cite{tau}, etc. To effectively achieve the fitting between predictions and true values, MSE loss and the AdamW optimizer are employed in this work. 

\textbf{Results and Analysis.} In this section, we primarily showcase the results derived from the Himawari-8/9 satellite sub-dataset. Additional results from other sub-datasets will be detailed in the Appendix~\ref{sec:trackc_details}. The quantitative results of the Track C spatio-temporal prediction are shown in Table.~\ref{tab:ST_prediction}. It can be drawn that our super data cubes are capable of supporting the selected methods for continuous prediction. Among them, recurrent-based PredRNN~\cite{predrnn} demonstrates outstanding stability across various evaluation metrics. Additionally, predicting across multiple regions is more challenging for the same model compared to single-region predictions, due to the more complex backgrounds in multi-region tasks. The qualitative results are provided in Fig.~\ref{Fig_result_prediction}. The models can predict the approximate locations of land, sea, and clouds. However, these methods still struggle with detail recovery and often produce blurred predictions. The inherent ambiguity of clouds and fog further exacerbates this issue. The underlying logic of cloud/fog discrimination in images relies on texture features, thus addressing the issue of blurriness in spatio-temporal prediction tasks is a key point for further research and reality application.


\section{Potential and limitations}
\label{sec:potential and limitations}

Potentially, we aspire for M4Fog to facilitate transfer learning or domain adaptation research to mitigate the discrepancies caused by different maritime regions, seasons, and satellite observation types.
By eliminating these variations, constructing a foundational model for global marine fog detection and forecasting can enhance its generalization capabilities worldwide. Furthermore, researchers can aggregate existing global weather or climate datasets to diversify fog-related services by additionally incorporating similar WeatherBench~\cite{weatherbench} dataset for visibility inversion and forecasting. This also enables leveraging specialized meteorological knowledge to optimize the marine fog task agent, adjusting from intensity levels of marine fog to visibility strength.

Nevertheless, M4Fog currently lacks multi-modal meteorological data for marine fog events during nighttime, dawn, and dusk, ensuring continuity only during daytime periods. To address the current limitations of M4Fog, we will extend its focus beyond daytime marine fog to achieve uninterrupted all-day detection and forecasting through integrating more extensive meteorological data sources. Moreover, crucial physical indicators for characterizing fog formation and dissipation, such as marine fog top height and optical thickness, will be included.

\section{Conclusions}

In this work, we introduce M4Fog, the most comprehensive global multi-regional, multi-modal, and multi-stage marine fog detection and forecasting dataset to date. It integrates meteorological data related to marine fog from meteorological satellites, observation stations, and numerical analyses over the past decade, and includes pixel-level annotations by meteorological experts. M4Fog covers the 15 major marine fog-prone regions globally and ensures the completeness of each marine fog event as much as possible. We define three fundamental tracks for M4Fog and conduct a series of experiments to benchmark existing ML methods and validate new baselines, laying a foundation for future research. All data and implementations are open-sourced to promote usage.

Our next focus for M4Fog is to supplement nighttime marine fog data and incorporate more forms of meteorological data to meet the requirements for around-the-clock marine fog detection and forecasting. We hope that M4Fog will foster exciting interdisciplinary collaboration between meteorology and artificial intelligence, enhance the understanding of marine fog formation and dissipation mechanisms, and improve the modeling of marine fog patterns, thereby developing more reliable systems for marine fog detection and forecasting, which are crucial for safe marine operations.

\begin{ack}
We sincerely thank our colleagues and collaborators for their invaluable support and insightful feedback throughout this research. We also express our gratitude to institutions and organizations, such as the China National Meteorological Centre (NMC), for providing the necessary resources and facilities. This study is supported by National Natural Science Foundation of China (NSFC) under Grant U23B2052. Additionally, this research is supported by Super Computing Platform of Beijing University of Posts and Telecommunications.
\end{ack}

\bibliographystyle{plainnat}
\normalem
\bibliography{refs} 


\clearpage
\appendix

\renewcommand{\thefigure}{A\arabic{figure}}
\setcounter{figure}{0}
\renewcommand{\thetable}{A\arabic{table}}
\setcounter{table}{0}

\section*{\centering Supplementary Materials for M4Fog}

\section{More Details of the Multi-modal Data in M4Fog}
\label{sec:data_details}

This section will introduce the multi-modal meteorological data related to marine fog used in M4Fog, which will be made public on the Github repository\footnote{The Github repository:~\url{https://github.com/kaka0910/M4Fog}} and project homepage\footnote{The project homepage address:~\url{https://kaka0910.github.io/}}. The introduction includes relevant information, collection and processing methods, and visualization results. For ease of reading, we have divided the data into two categories: dense and sparse, and will introduce each category individually. It is important to note that all times referenced in this paper are in Coordinated Universal Time (UTC).


\subsection{Dense Data}

\subsubsection{Satellite Remote Sensing Data}

\begin{table}[h]
\centering
\caption{Center Wavelengths (CW) and Spatial Resolutions (SR) of typical imagers from four series of geostationary satellite (FengYun-4, Himawari-8/9, GOES-16, Meteosat-11) used in M4Fog.}
\label{tab:satellites}
\resizebox{\textwidth}{!}{%
\begin{tabular}{ll|cccc|cccc|ccccccccccc|c}
    \toprule
     \multicolumn{2}{l|}{\textbf{Satellites}} & \multicolumn{4}{c|}{\textbf{Visible}} & \multicolumn{4}{c|}{\textbf{Near-Infrared}} & \multicolumn{11}{c|}{\textbf{Infrared}} & \textbf{Nums.} \\
    
    \midrule
    \multirow{2}{*}{\textbf{FY4A}}  & Index: & Band01 & & Band02 & & Band03 & Band04 & Band05 & Band06 & \multicolumn{2}{|l}{Band07/08}  & Band09 & & Band10 & Band11 & & Band12 & & Band13 & Band14 & \multirow{3}{*}{14} \\
    & CW($\mu$m): & 0.47 & & 0.65 & & 0.825 & 1.375 & 1.61 & 2.25 & 3.75 & & 6.25 & & 7.1 & 8.5 & & 10.7 & & 12.0 & 13.5 &  \\
    & SR(km): & 1 & & 0.5/1 & & 1 & 2 & 2 & 2/4 & 4 & & 4 & & 4 & 4 & & 4 & & 4 & 4 &  \\
    
    \midrule
    \multirow{2}{*}{\textbf{H8/9}}  & Index: & Band01 & Band02 & Band03 & & Band04 & & Band05 & Band06 & & Band07 & Band08 & Band09 & Band10 & Band11 & Band12 & Band13 & Band14 & Band15 & Band16 & \multirow{3}{*}{16} \\
    & CW($\mu$m): & 0.47 & 0.51 & 0.64 & & 0.86 & & 1.6 & 2.3 & & 3.9 & 6.2 & 6.9 & 7.3 & 8.6 & 9.6 & 10.4 & 11.2 & 12.4 & 13.3 & \\
    & SR(km): & 1 & 1 & 0.5 & & 1 & & 2 & 2 & & 2 & 2 & 2 & 2 & 2 & 2 & 2 & 2 & 2 & 2 &  \\

    \midrule
    \multirow{2}{*}{\textbf{GOES16}}  & Index: & Band01 & & Band02 & & Band03 & Band04 & Band05 & Band06 & & Band07 & Band08 & Band09 & Band10 & Band11 & Band12 & Band13 & Band14 & Band15 & Band16 & \multirow{3}{*}{16} \\
    & CW($\mu$m): & 0.47 & & 0.64 & & 0.86 & 1.38 & 1.61 & 2.26 & & 3.90 & 6.15 & 7.00 & 7.40 & 8.50 & 9.70 & 10.30 & 11.20 & 12.30 & 13.3 & \\
    & SR(km): & 1 & & 0.5 & & 1 & 2 & 1 & 2 & & 2 & 2 & 2 & 2 & 2 & 2 & 2 & 2 & 2 & 2 &  \\

    \midrule
    \multirow{2}{*}{\textbf{Meteo11}} & Index: & & & Band01 & Band02 & Band03 & & Band04 & & & Band05 & Band06 & & Band07 & Band08 & Band09 & Band10 & & Band11 & Band12 & \multirow{3}{*}{12} \\
    & CW($\mu$m): & & & 0.635 & 0.75 & 0.81 & & 1.64 & & & 3.92 & 6.25 & & 7.35 & 8.70 & 9.66 & 10.80 & & 12.00 & 13.40 & \\
    & SR(km): & & & 3 & 1 & 3 & & 3 & & & 3 & 3 & & 3 & 3 & 3 & 3 & & 3 & 3 &  \\
    \bottomrule
\end{tabular}%
}
\end{table}

The satellite remote sensing data in M4Fog are sourced from four series of geostationary meteorological satellites, provided by various countries and regions around the world. These satellites are a crucial component of the Satellite Observations in the Global Observing System (GOS)\footnote{The introduction of Global Observing System (GOS) with the official website~\url{https://wmo.int/activities/global-observing-system-gos/global-observing-system-gos}.}. Table.~\ref{tab:satellites} presents some key parameters of these satellites that influence our data processing, including the center wavelengths and spatial resolutions of each band. These satellites orbit at an altitude of approximately 38,500 kilometers above the Earth's equator, matching the Earth's rotation and making them appear stationary at a fixed point above the equator (known as the sub-satellite point). Because meteorological geostationary satellites can continuously observe specific regions, covering approximately 42\% of the Earth's surface, they are considered dense data sources. Next, we will further introduce each series and how we utilize them.


\begin{figure}[ht] 
\centering 
\includegraphics[width=\textwidth]{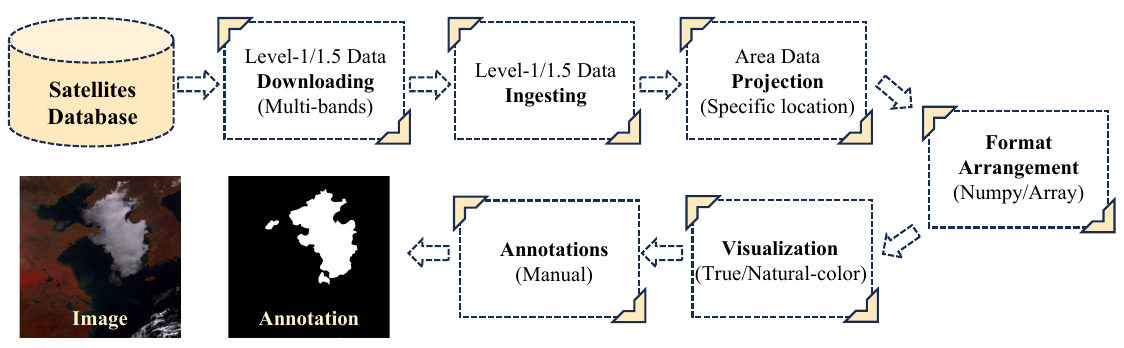} 
\caption{Basic processing workflow for Level-1 Data applicable to various satellites, including L1 data downloading, ingesting, projection, format arrangement, visualization and annotation.} 
\label{fig:satellite_flowchart} 
\end{figure}

\textbf{FengYun-4 (FY4) Series of Satellites.} The FengYun-4 (FY4) series of geostationary satellites is a critical component of China's Fengyun meteorological satellite program,  launched and operated by the National Satellite Meteorological Center of the China Meteorological Administration. The FY4 series satellites produce Level 1 (L1) and Level 2 (L2) data products, providing foundational observation data and advanced analysis products, respectively. Among them, L1 data are foundational data products obtained after geometric and radiometric corrections of raw observation data, including Brightness Temperature and Spectral Radiance. FY4A, the first satellite in the series, is equipped with advanced imaging instruments such as the Advanced Geosynchronous Radiation Imager (AGRI), the Geostationary Interferometric Infrared Sounder (GIIRS) and other imagers. It can provide data with a temporal resolution of 15 minutes, providing high-resolution imaging and precise monitoring of the Earth's atmosphere, oceans and land. As of 2024, the FY4 series primarily includes the FY4A and FY4B geostationary satellites\footnote{According to the announcement from the National Satellite Meteorological Center, the FY4A satellite temporarily ceased meteorological services on May 23, 2024.}. 

\begin{figure}[ht] 
\centering 
\includegraphics[width=\textwidth]{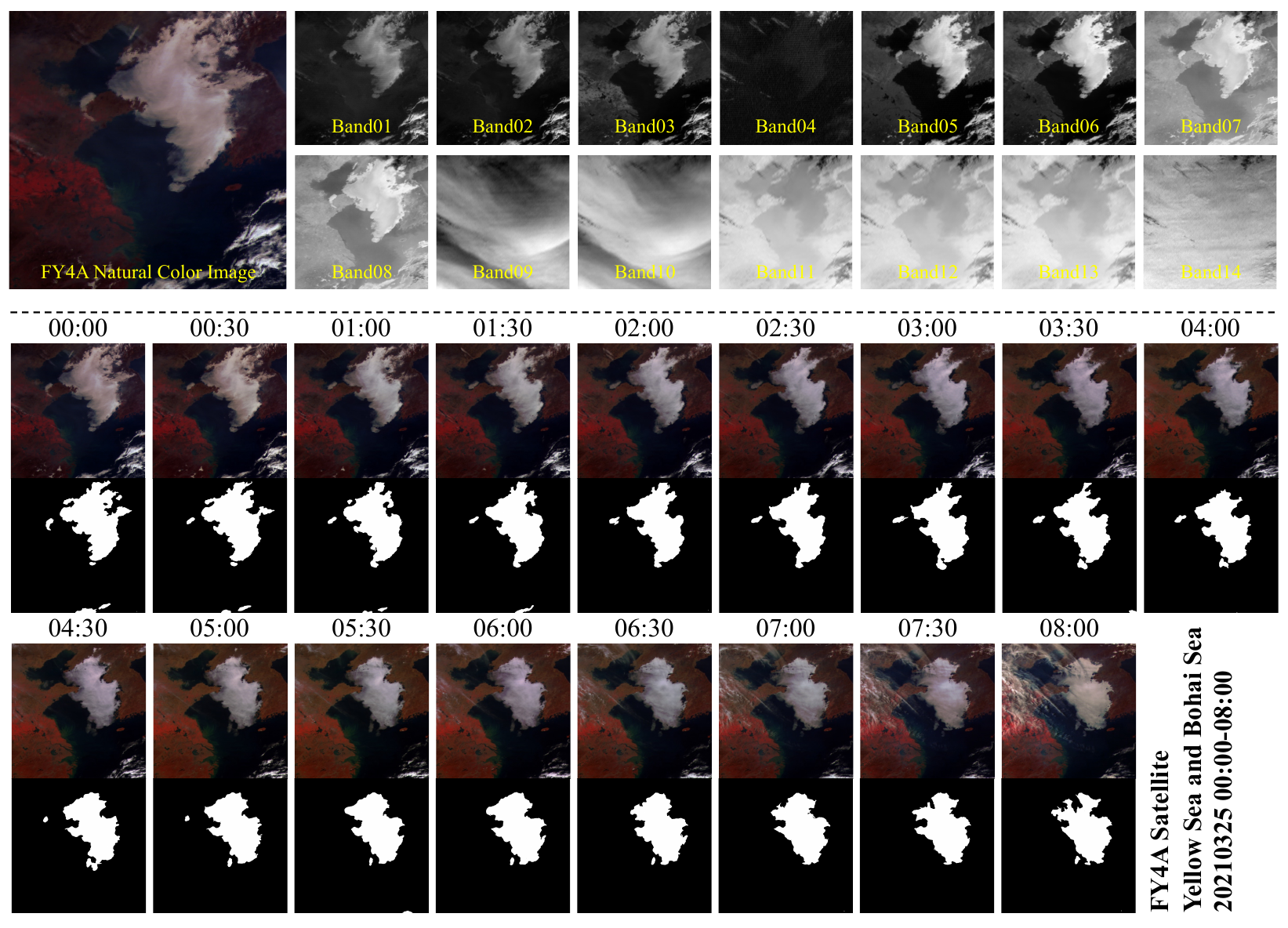} 
\caption{Visualization examples over the Yellow and Bohai Seas, (a) natural-color image with different visualized band images taking 00:00, March 25, 2021 as an example, (b) a super data cube from the FY4A satellite along with manually labels on March 25, 2021, from 00:00 to 08:00.} 
\label{fig:satellite_fy4} 
\end{figure}

In this study, we utilize L1 data from all 14 bands of the FY4A-AGRI related to marine fog detection and forecasting. Fig.~\ref{fig:satellite_flowchart} illustrates the basic processing workflow for the data, applicable to other satellites. First, leveraging marine weather reviews published by meteorological centers, selected publicly available L1 raw data is downloaded from the Fengyun Satellite Data Center\footnote{The website for downloading FY4 data:~\url{http://satellite.nsmc.org.cn/PortalSite/}}. The raw data, within specific latitude and longitude ranges, is then obtained after projection and is typically available in widely distributed decimal format. These raw data are sorted by date, region, and channel, and consolidated into a Numpy file suitable for a machine learning framework. Next, visual images are generated through true or natural color synthesis and manually annotated by meteorology experts. The process for synthesizing FY4A data into natural-color images is detailed as follows: data $x_i \in \mathbb{R}^{H \times W}$ selected from band 01 (0.47$\mu$m), band 02 (0.65$\mu$m), and band 03 (0.825$\mu$m) is scaled to the 0-255 range to form new data $x'_i$ of the same shape.  An example of the proposed super data cube along with manually annotated labels is illustrated in Fig.~\ref{fig:satellite_fy4}, showing data collected from the FY4A satellite over the Yellow and Bohai Seas on March 25, 2021, from 00:00 to 08:00.



\textbf{Himawari8/9 (H8/9) Series of Satellites.} The Himawari 8/9 series geostationary satellites, operated by the Japan Meteorological Agency (JMA)\footnote{The website of JAXA Himawari Monitor for H8/9 data downloading:~\url{https://www.eorc.jaxa.jp/ptree/index.html}}, are critical for weather monitoring and environmental observation in the Asia-Pacific region, delivering real-time data every 10 minutes. These satellites feature the Advanced Himawari Imager (AHI), which provides high-resolution, multi-spectral imagery across 16 channels, covering visible, near-infrared, and infrared bands. The data from the Himawari-9 satellite has been used by JAXA since December 13, 2022\footnote{The Notification of Satellite Transition:~\url{https://www.eorc.jaxa.jp/ptree/userguide.html}}, following the switchover from Himawari-8. This transition ensures the continuity of high-resolution geostationary weather observation over the Asia-Pacific region.

\begin{figure}[ht] 
\centering 
\includegraphics[width=\textwidth]{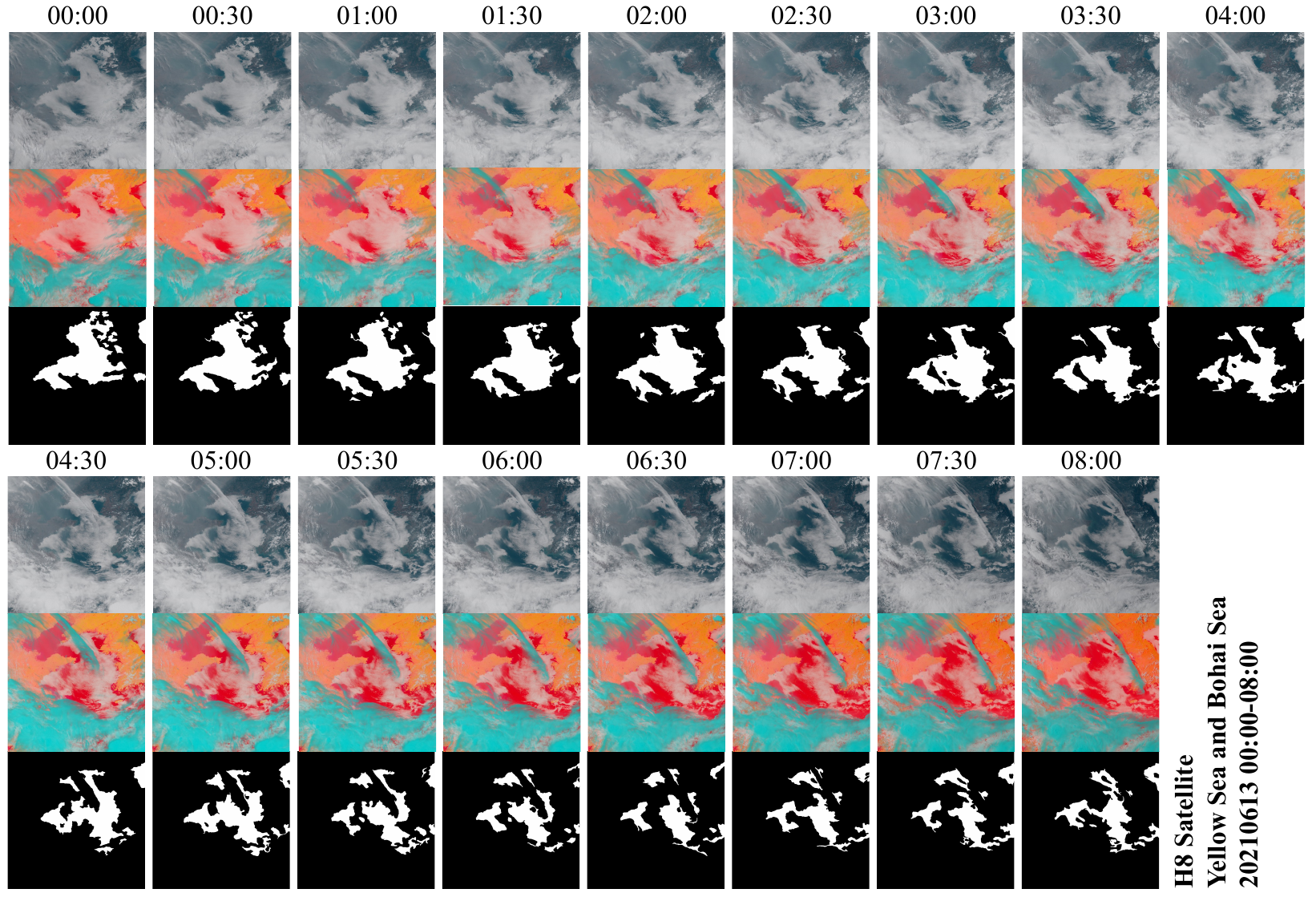} 
\caption{Visualization example of a super data cube from the H8 satellite along with manually annotated labels based on true-color and natural-color imageries, over the Yellow and Bohai Seas on June 13, 2021, from 00:00 to 08:00.} 
\label{fig:satellite_h8} 
\end{figure}

Similar to the processing of FY4 series of satellite data, the L1 data from all 16 bands of the H-8/9 is also selected. The data is processed step by step, including downloading, ingesting, projection, format arrangement, visualization and annotation. In addition to using Band 01 (0.47$\mu$m), Band 02 (0.51$\mu$m), and Band 03 (0.64$\mu$m) to create true-color images from H8, we also provide a method for synthesizing natural-color images. According to the paper, we use Band 03 (0.64$\mu$m), Band 04 (0.86$\mu$m), and Band 14 (11.20$\mu$m) to better highlight the features and textures of marine fog, relying on the high reflectance and low brightness temperature values of the marine fog. Specifically, Bands 01, 02, 03, and 04 are directly scaled to the 0-255 range, while Band 14 is adjusted to the image value range by subtracting 200. It is worth noting that this method of synthesizing natural-color images is also applicable to other series of satellites, as long as the band center wavelengths are similar. The Fig.~\ref{fig:satellite_h8} displays the super data cubes and corresponding annotations created for 17 consecutive timestamps with a half-hour interval from H-8/9 satellites in M4Fog, over the Yellow and Bohai Seas on June 13, 2021, from 00:00 to 08:00.


\textbf{GOES (GOES16) Series of Satellites.} The Geostationary Operational Environmental Satellites (GOES) series, operated by the National Oceanic and Atmospheric Administration (NOAA)\footnote{The website of GOES:~\url{https://www.goes.noaa.gov/} and NOAA-GOES Imagery Viewer:~\url{https://www.star.nesdis.noaa.gov/goes/index.php}}, play a vital role in weather forecasting, severe storm tracking, and environmental monitoring in the Western Hemisphere. The latest satellites, GOES-16 (GOES-East) and GOES-17 (GOES-West), are equipped with advanced instruments such as the Advanced Baseline Imager (ABI), which provides high-resolution imagery across 16 spectral bands. These satellites offer rapid scanning capabilities, providing real-time data every 5 minutes for full-disk images and as often as every 30 seconds for specific regions, significantly enhancing the accuracy and timeliness of meteorological services. In particular, GOES-16\footnote{The data and products downloading access through Geostationary Operational Environmental Satellites-R series data products:~\url{https://www.goes-r.gov/products/overview.html}}, delivers Level 1 data, including calibrated radiance and lightning detection, and Level 2 products such as cloud properties, sea and land surface temperatures, aerosol data and others products.


\begin{figure}[ht] 
\centering 
\includegraphics[width=\textwidth]{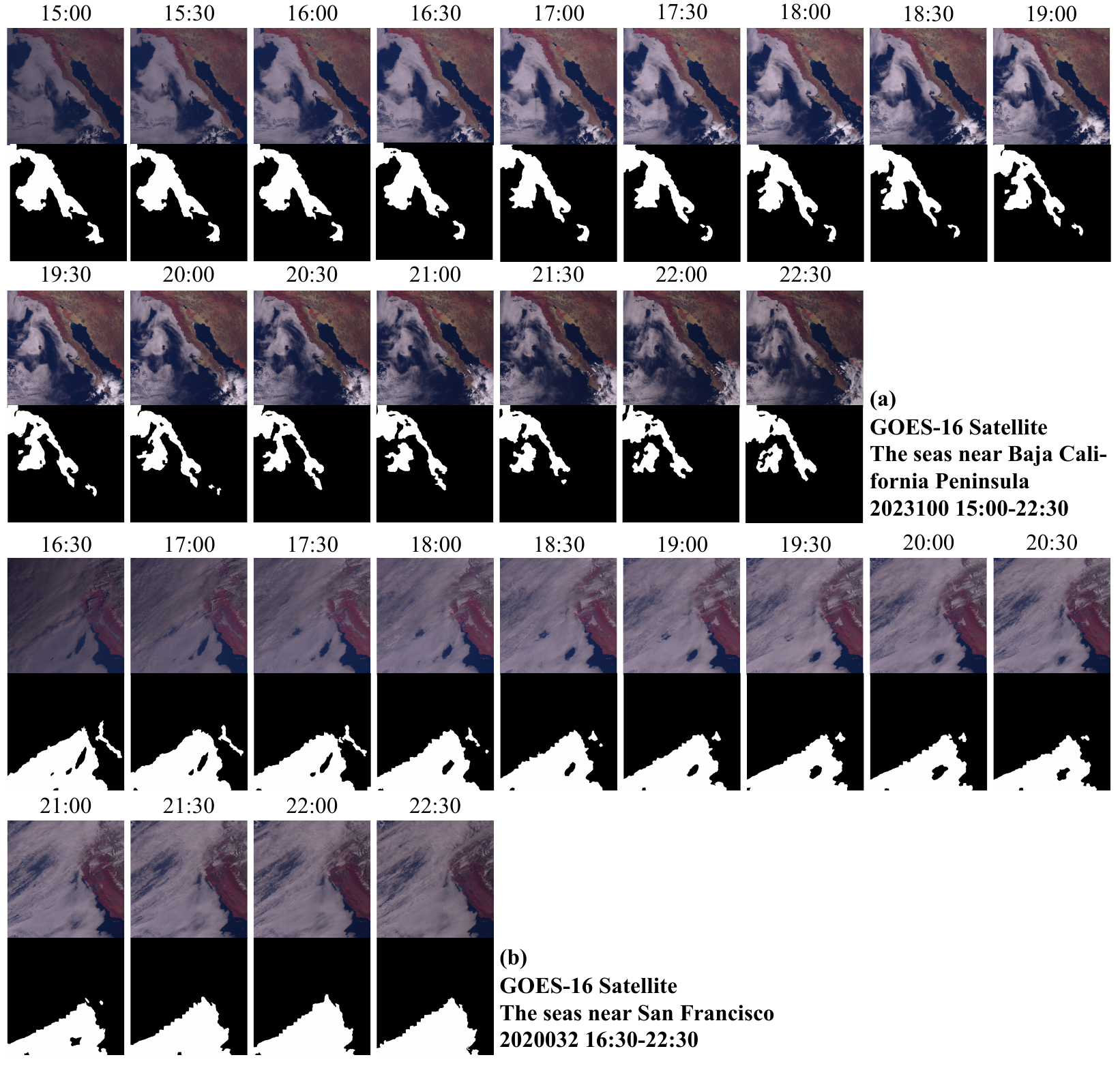} 
\caption{Visualization examples of two super data cubes from the GOES-16 satellite along with manually annotated labels, over (a) the seas near Baja California Peninsula on 100rd (April 10), 2023, from 15:00 to 22:30, (b) the seas near San Francisco on 32rd (February 1), 2020, from 16:30 to 22:30.} 
\label{fig:satellite_goes} 
\end{figure}

In this work, we selected the GOES-16 satellite as one of the data foundations for the super data cubes over the Western Hemisphere. The L1 data with all 16 bands are utilized. To maintain a similar temporal resolution to other satellites, we chose data at half-hour intervals. It is important to note that the time span varies by region due to the influence of solar radiation on visible light imaging band data. In other words, the higher the latitude of the region, the fewer data sequences there are in the super data cubes. For example, the time range for the seas near San Francisco is from 16:30 to 22:30, while for the Baja California Peninsula, it is from 15:00 to 22:30. Visualization examples are shown in Fig.~\ref{fig:satellite_goes}.

\textbf{MeteoSat (Meteo11) Series of Satellites.} The Meteosat Second Generation (MSG) satellites\footnote{The details of Meteosat Second Generation (MSG):~\url{https://user.eumetsat.int/data/satellites/meteosat-second-generation}}, operated by the European Organisation for the Exploitation of Meteorological Satellites (EUMETSAT), are designed to enhance weather forecasting and climate monitoring. Launched from 2002 onwards, including MeteoSat-8 to MeteoSat-11, these satellites provide high-resolution, multi-spectral images every 15 minutes. Equipped with the Spinning Enhanced Visible and InfraRed Imager (SEVIRI), which has 12 spectral channels covering visible, near-infrared, and infrared bands, MSG satellites enable detailed monitoring of cloud structures, sea surface temperatures, atmospheric water vapor, and other vital meteorological parameters. These capabilities significantly improve weather forecasts, support long-term climate research, and aid in environmental monitoring and disaster response.

\begin{figure}[ht] 
\centering 
\includegraphics[width=\textwidth]{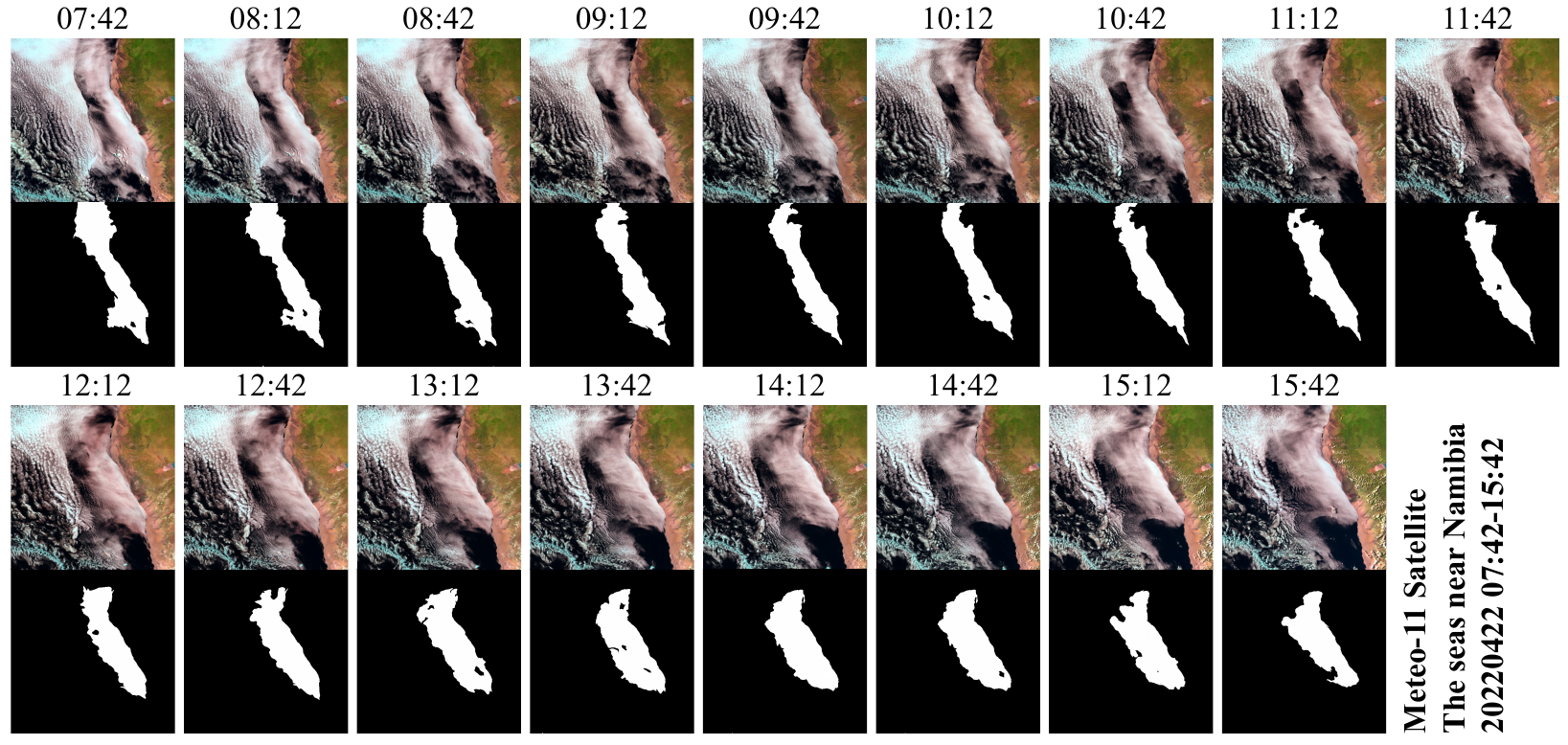} 
\caption{Visualization example of a super data cube from the MeteoSat-11 satellite along with manually annotated labels based on natural-color images, over the seas near Namibia in Africa on April 22, 2022, from 07:42 to 15:42.} 
\label{fig:satellite_meteo} 
\end{figure}

Unlike the other satellites that use Level-1 data, High Rate SEVIRI Level 1.5 (L1.5) Image Data\footnote{The Level 1.5 Image Data address:~\url{https://user.eumetsat.int/catalogue/EO:EUM:DAT:MSG:HRSEVIRI/overview}}, with half-hour intervals, is specifically utilized in M4Fog. The data processing workflow is consistent with the L1 data. L1.5 data is satellite remote sensing data positioned between L1 and L2, undergoing geometric and radiometric corrections, as well as georeferencing. It retains the original observational information, making it suitable for applications that require high precision and further processing, such as weather monitoring and environmental surveillance. Following to SatPy\footnote{The homepage of SatPy Python package~\url{https://satpy.readthedocs.io/en/stable/overview.html} and Github repository~\url{https://github.com/pytroll/satpy/tree/main}} using NIR (1.64$\mu$m), VIS (0.81$\mu$m) and VIS (0.635$\mu$m), the natural-color images are synthesized that a visualization example of super data cube collted from MeteoSat is shown in the Fig.~\ref{fig:satellite_meteo} over the seas near Namibia. Among them, whitish clouds represent low-level clouds, including low clouds and fog, while approximately bright blue areas indicate higher-altitude clouds.

\subsubsection{Numerical Analysis Data} 

Sea Surface Temperature (SST) refers to the temperature of the water at the ocean's surface, typically measured within the upper few centimeters using satellite radiometers, buoys, and ships. Accurate SST measurements are crucial for understanding the heat exchange between the ocean and atmosphere, predicting weather and climate changes, and monitoring marine ecosystems. SST significantly affects the formation of sea fog, as warm, moist air cools and condenses into fog when it flows over colder sea surfaces. Specifically, from the weather phenomenon events in the paper~\cite{sst_cases}, it can be seen that warm and humid air flows from areas of warmer SST to areas of colder SST along the coast to form sea fog. Additionally, early study~\cite{sst_front} focuses on the effects of background circulation and Sea Surface Temperature Front (SSTF) on the transition of stratus clouds into sea fog. Therefore, incorporating SST data into our M4Fog is essential for more accurately predicting and analyzing the occurrence of marine fog.

\begin{figure}[h] 
\centering 
\includegraphics[width=\textwidth]{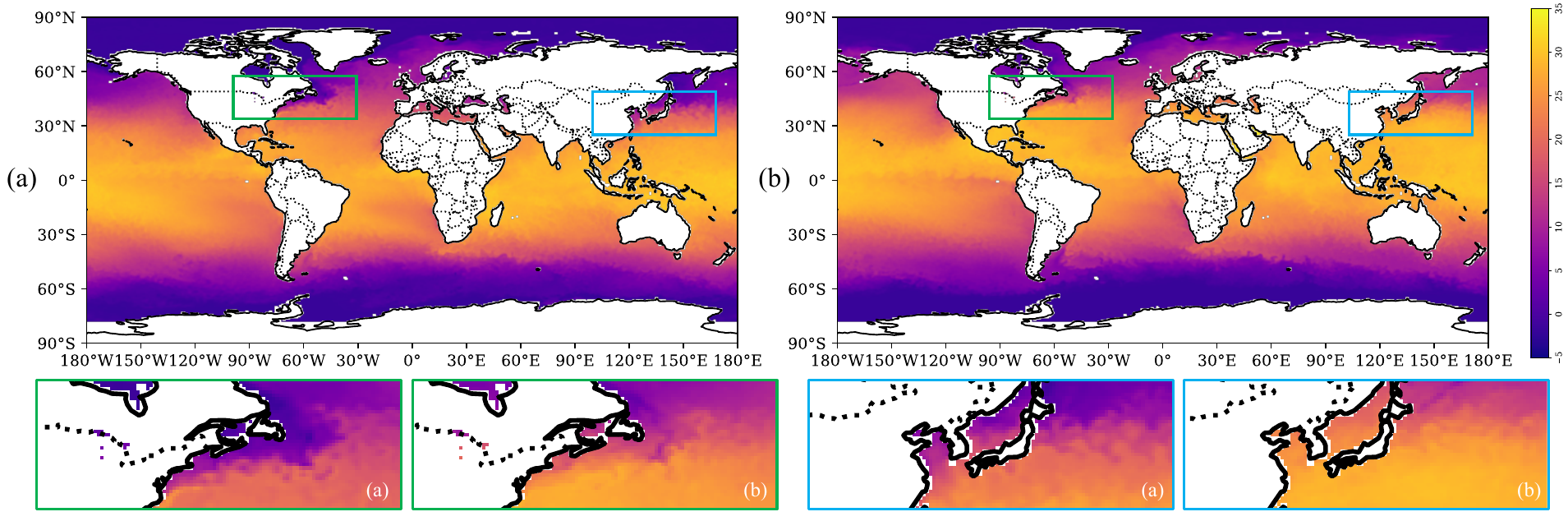} 
\caption{Visualization examples of global Sea Surface Temperatures (SST) on January 1 and April 10, 2018, highlighting several areas of interest in M4Fog.} 
\label{fig:sst} 
\end{figure}

Currently, SST data is available from various sources, including NOAA, EUMETSAT, JAXA, and others. These platforms provide high-resolution, real-time, and historical SST data for weather forecasting, climate research, and ocean monitoring. Due to the high resolution and real-time updates of NOAA's SST data, we selected the NOAA OI SST V2 High Resolution Dataset\footnote{The address of NOAA-SST Data:~\url{https://psl.noaa.gov/data/gridded/data.noaa.oisst.v2.highres.html}} for our study. In Fig.~\ref{fig:sst}, we present visualization examples of global sea surface temperatures for two days, January 1 and April 10, 2018. The figure also highlights several sea areas of interest within M4Fog.

Additionally, due to the differences in surface reflectance, temperature, and radiation characteristics between ocean and land, we introduced a corresponding sea-land mask for every sea areas in the construction of the M4Fog dataset. This helps reduce errors and achieve higher accuracy in marine fog detection and forecasting, facilitating more precise identification and analysis of marine fog formation and distribution.

\subsection{Sparse Data}

\subsubsection{ICOADS Data} 

The International Comprehensive Ocean-Atmosphere Data Set (ICOADS)\footnote{The access of downloading ICOADS data:~\url{https://rda.ucar.edu/datasets/ds548.0/dataaccess/}} is a global dataset of marine and atmospheric observations spanning from 1662 to the present. It integrates data from various sources, including ships, buoys, marine platforms, and satellites, covering key parameters such as sea surface temperature, air pressure, wind speed, and humidity. ICOADS is widely used in climate research, weather forecasting, oceanography, and environmental monitoring.

\begin{table}[ht]
\centering
\caption{Some samples of selected ICOADS records including timestamp, location and fog-related meteorological elements, with labels indicating the presence or absence of marine fog during 2021 over Yellow and Bohai Seas.}
\label{tab:icoads}
\resizebox{\textwidth}{!}{%
\begin{tabular}{c|cccccccccc}
    \toprule
    Index & Year & Month & Day & Hour & Minute & Longitude & Latitude & Visibility & Present Weather & Label \\
    \midrule
    1 & 2021 & 1 & 24 & 8 & 0 & 120.1 & 35.4 & 94.0 & 45.0 & \cellcolor{red!25}Fog \\
    2 & 2021 & 2 & 20 & 7 & 0 & 121.4 & 40.1 & \textcolor{green}{96.0} & 5.0 & \cellcolor{green!25}Unfog \\
    3 & 2021 & 3 &  4 & 6 & 0 & 122.5 & 30.4 & \textcolor{green}{96.0} & 5.0 & \cellcolor{green!25}Unfog \\
    4 & 2021 & 4 & 28 & 0 & 0 & 122.4 & 31.3 & 91.0 & 47.0 & \cellcolor{red!25}Fog \\
    5 & 2021 & 5 & 20 & 3 & 0 & 122.8 & 29.8 & 94.0 & \textcolor{green}{63.0} & \cellcolor{green!25}Unfog \\
    6 & 2021 & 6 & 13 & 8 & 0 & 122.0 & 38.8 & 94.0 & 45.0 & \cellcolor{red!25}Fog \\
    7 & 2021 & 7 & 11 & 4 & 0 & 126.2 & 32.9 & \textcolor{green}{96.0} & 2.0 & \cellcolor{green!25}Unfog \\
    8 & 2021 & 8 & 10 & 6 & 0 & 125.5 & 32.0 & \textcolor{green}{98.0} & 2.0 & \cellcolor{green!25}Unfog \\
    \bottomrule
\end{tabular}%
}
\end{table}

In our study, ICOADS data serves two main purposes. First, it provides a basis for analyzing the regions and seasons of global marine fog events from 2015-2024, aiding in the selection of typical marine fog events for M4Fog. Second, it offers a sparse validation metric for Static and Dynamic marine fog detection. As mentioned in main text, the criteria for marine fog determination are $VV \leq 94, WW<60$\footnote{The introduction of related meteorological elements in ICOADS:~\url{https://rda.ucar.edu/OS/web/datasets/ds548.0/docs/R3.0.2-imma1_short.pdf}}, where VV and WW are visibility and the present weather, respectively. The Table.~\ref{tab:icoads} below presents some ICOADS records along with their respective labels indicating the presence or absence of marine fog. Additionally, green text indicates non-compliant meteorological elements identified as marine fog. 

\subsubsection{Coastal Observation Station Data} 

In addition to the sparse marine navigation data from ICOADS, the National Meteorological Center of China has provided meteorological observation data from coastal stations along the Yellow and Bohai Seas for M4Fog. These stations record wind speed, humidity, temperature, and other relevant meteorological variables at half-hour intervals, which can assist in determining sea fog conditions. Unlike ICOADS, which uses number codes to represent certain meteorological variables, the coastal station data provides the actual values of these elements. 

Although the official criterion for determining marine fog is a horizontal visibility of less than 1 kilometer, during the comparison of observational data with satellite data, we found that observations with visibility between 1 and 2 kilometers were very similar to those with visibility less than 1 kilometer. As a result, we expanded our marine fog criteria to include visibility of less than 2 kilometers. Due to the confidentiality and sensitivity of these meteorological data, we will not release the raw data publicly. However, a public testing platform is available on our homepage.


\section{More Details of the Multi-regional and Multi-stage Data in M4Fog}
\label{sec:analysis_season}

\begin{figure}[ht] 
\centering 
\includegraphics[width=\textwidth]{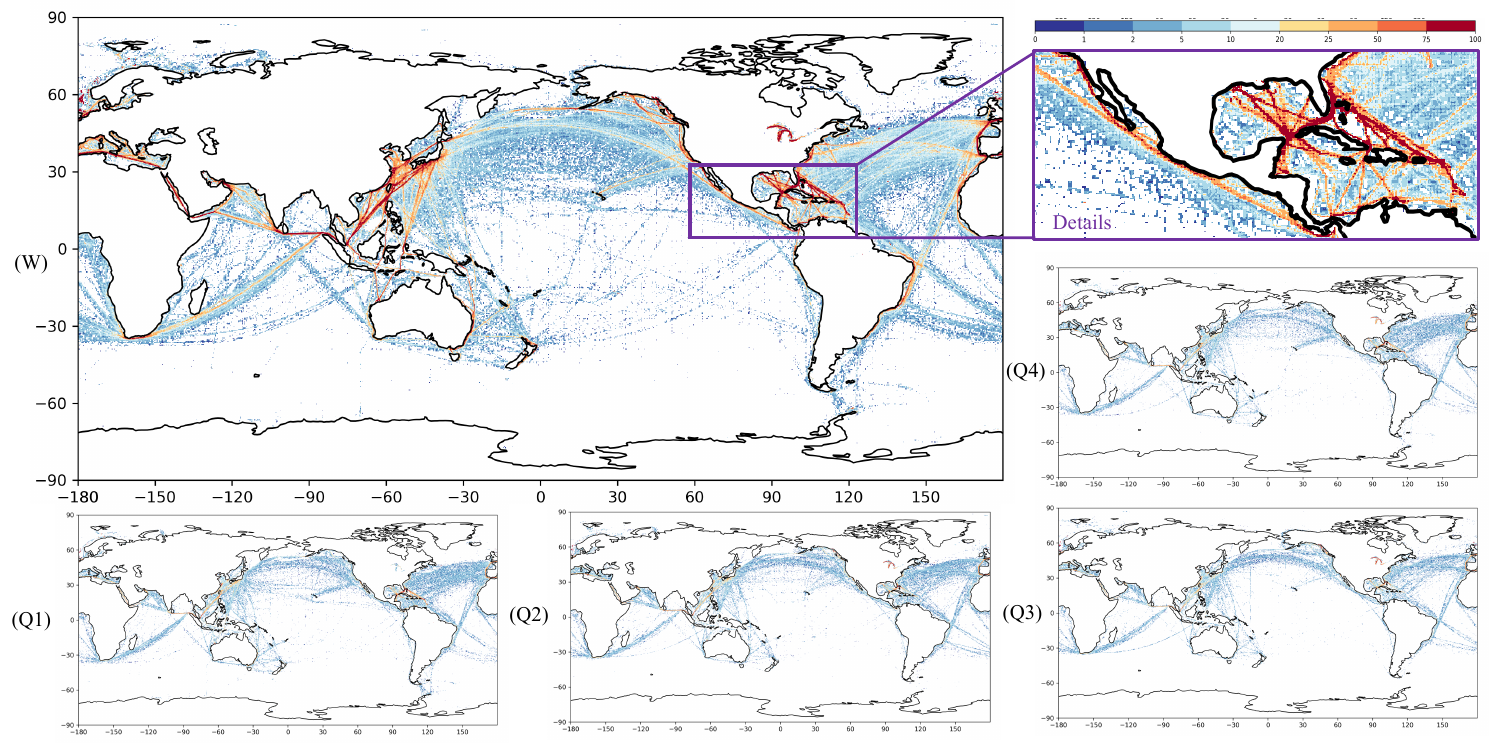} 
\caption{Global marine fog frequency map (W) based on ICOADS and fog-related navigation records from 2015-2024, with quarterly subplots (Q1-Q4).} 
\label{fig:Fig_season} 
\end{figure}

By screening ICOADS and fog-related navigation records from 2015-2024, we created a global marine fog frequency map based on the cumulative frequency of fog occurrences, as shown in Fig.~\ref{fig:Fig_season} (W). The frequency map is further divided into subplots for each of the four quarters, detailed in Fig.~\ref{fig:Fig_season} (Q1)-(Q4). The comparison indicates that marine fog is not only a global meteorological phenomenon but also exhibits different seasonal stage variations across various sea areas. These also are crucial factors in selecting marine fog events for M4Fog. 

\begin{table}[ht]
\centering
\caption{Detailed information for Multi-regional proposed in M4Fog, including Regions designation, Abbreviation (regions abbr.-Continent/Ocean abbr.), Location, Supported satellite data, and corresponding Major City/Port.}
\label{tab:regions}
\resizebox{\textwidth}{!}{%
\begin{tabular}{l|c|cc|cccc|c}
    \toprule
    \multirow{2}{*}{\textbf{Regions}} & \multirow{2}{*}{\textbf{Abbr.}} & \multicolumn{2}{c|}{\textbf{Locations}} & \multicolumn{4}{c|}{\textbf{Supported Satellites}} & \multirow{2}{*}{\textbf{Major City/Port}} \\
    & & Latitude range & Longitude range & FY4 & H8/9 & MSG & GOES & \\
    
    \midrule
    Yellow and Bohai Seas  & YB, AS/PO & 28.7$^\circ$N-41.5$^\circ$N & 116.2$^\circ$E-129.0$^\circ$E & \ding{51} & \ding{51} & & & Tianjin \\
    China East Sea         & CE, AS/PO & 20.0$^\circ$N-32.8$^\circ$N & 117.0$^\circ$E-129.8$^\circ$E & \ding{51} & \ding{51} & & & Shanghai \\
    China North Sea        & CN, AS/PO & 11.2$^\circ$N-24.0$^\circ$N & 105.0$^\circ$E-117.8$^\circ$E & \ding{51} & \ding{51} & & & Hong Kong \\
    
    \midrule
    Mediterranean (East)   & ME, EU/AO & 37.2$^\circ$N-50.0$^\circ$N & 27.0$^\circ$E-39.8$^\circ$E & & & \ding{51} & & Athens \\
    Mediterranean (Central) & MC, EU/AO & 30.0$^\circ$N-42.8$^\circ$N & 15.0$^\circ$E-27.8$^\circ$E & & & \ding{51} & & Rome \\
    Mediterranean (West)   & MW, EU/AO & 33.0$^\circ$N-45.8$^\circ$N & 0.0$^\circ$-12.8$^\circ$E & & & \ding{51} & & Barcelona \\
    North Sea              & NS, EU/AO & 47.2$^\circ$N-60.0$^\circ$N & 7.8$^\circ$W-5.0$^\circ$E & & & \ding{51} & & Rotterdam \\
    Namibia                & BE, AF/IO & 14.0$^\circ$S-26.8$^\circ$S & 2.0$^\circ$E-16.8$^\circ$E & & & \ding{51} & & Cape Town \\
    Agulhas Current        & AG, AF/IO & 25.2$^\circ$S-38.0$^\circ$S & 8.0$^\circ$E-20.8$^\circ$E & & & \ding{51} & & Durban \\
    
    \midrule
    Gulf of Alaska         & GA, NA/PO & 42.2$^\circ$N-55.0$^\circ$N & 120.0$^\circ$W-132.8$^\circ$W & & & & \ding{51} & Anchorage \\
    California Current     & CC, NA/PO & 32.0$^\circ$N-44.8$^\circ$N & 119.0$^\circ$W-131.8$^\circ$W & & & & \ding{51} & San Francisco \\
    Baja California        & BC, NA/PO & 22.2$^\circ$N-35.0$^\circ$N & 109.0$^\circ$W-121.8$^\circ$W & & & & \ding{51} & La Paz \\
    Gulf Stream (N. Atlantic) & GS, NA/AO & 42.0$^\circ$N-54.8$^\circ$N &  57.0$^\circ$W-69.8$^\circ$W  & & & & \ding{51} & New York \\
    Gulf of Mexico         & GM, NA/GO & 18.0$^\circ$N-30.8$^\circ$N & 87.2$^\circ$W-100.0$^\circ$W  & & & & \ding{51} & New Orleans \\
    North Brazil Current   & NB, SA/AO & 18.0$^\circ$S-30.8$^\circ$S &  37.0$^\circ$W-49.8$^\circ$W  & & & & \ding{51} & Belém \\
      
    \bottomrule
\end{tabular}%
}
\end{table}

The detailed information of the global Multi-regional areas proposed in M4Fog is presented in Table.~\ref{tab:regions}, spanning multiple continents and oceans, and covering several important coastal cities and ports around the world. The abbreviations of Continents or Oceans\footnote{The abbreviations of Continents are: Asia (AS), Africa (AF), North America (NA), South America (SA), Antarctica (AN), Europe (EU), Australia (AU). The abbreviations of Oceans are: Pacific Ocean (PO), Atlantic Ocean (AO), Indian Ocean (IO), Southern Ocean (SO), Arctic Ocean (ArO).} are provided below. Due to the proximity of the sub-satellite points of the FY4 series and H8/9 series satellites\footnote{The FY4 series satellites have a sub-satellite point at approximately 104.7°E, while the H8/9 series satellites have a sub-satellite point at approximately 140.7°E.}, both can collaboratively monitor most regions of the Western Pacific. This collaboration allows for mutual supplementation and significantly enhances detection capabilities.

\section{More Details for Track A: Static Marine Fog Detection}
\label{sec:tracka_details}

\subsection{Setup}
The data composition of each sub-dataset in Track A Static Marine Fog Detection is shown in Table.~\ref{tab:tracka_subdatasets} based on different satellites. In this study, the all bands collected from satellite data are used to provide more comprehensive fog-related features.

\begin{table}[t]
\centering
\caption{The details of sub-datasets constructed for Track A Static Marine Fog Detection.}
\label{tab:tracka_subdatasets}
\tiny
\resizebox{\textwidth}{!}{%
\begin{tabular}{c|c|cc|c|ccc}
    \toprule
    \multirow{2}{*}{\textbf{Sub-datasets}} & \multirow{2}{*}{\textbf{Data Size}} & \multicolumn{2}{c|}{\textbf{Temporal Range}} & \multirow{2}{*}{\textbf{Regional Information}} & \multicolumn{3}{c}{\textbf{The Number of samples}} \\
    & & Range & Split & & Total & Training & Test \\
    
    \midrule
    \textbf{Himawari-8/9} & (1024, 1024, 16) & 2018-2021 & 2021 & Specific (Yellow/Bohai Seas) & 1,802  & 1,122  & 680 \\
    \textbf{FengYun-4A}   & (1024, 1024, 14) & 2018-2021 & 2021 & Specific (Yellow/Bohai Seas) & 1,724  & 1,054  & 670 \\
    
    \textbf{GOES-16}      & (1024, 1024, 16) & 2020-2023 & 2023 & Multi-regional (4 areas)     & 2,240  & 1,801  & 439 \\
    \textbf{MeteoSat-11}  & (1024, 1024, 12) & 2020-2022 & 2022 & Multi-regional (5 areas)     & 4,012  & 2,380  & 1,632 \\
      
    \bottomrule
\end{tabular}%
}
\end{table}

\subsection{Evaluation Metrics}

\subsubsection{Dense Metrics} 
Dense metrics are calculated for the entire predicted image based on manually annotated masks, with the specific calculation formulas as follows:

The Critical Success Index~(CSI):
\begin{equation}
\text{CSI} = \frac{TP_1}{TP_1 + FN_1 + FP_1} \tag{A1}
\end{equation}

The Accuracy (Acc): 
\begin{equation}
\text{Acc} = \frac{TP_1 + TN_1}{TP_1 + TN_1 + FP_1 + FN_1} \tag{A2}
\end{equation}

The mean Intersection over Union~(mIoU):
\begin{equation}
\text{mIoU} = \frac{1}{2} \sum_{i=0}^1 \frac{{TP}_i}{{TP}_i + {FP}_i + {FN}_i} \tag{A3}
\end{equation}

The mean Accuracy (Acc):
\begin{equation}
\text{mAcc} = \frac{1}{2} \sum_{i=0}^1 \frac{TP_i + TN_i}{TP_i + TN_i + FP_i + FN_i} \tag{A4}
\end{equation}

\subsubsection{Sparse Metrics} 
Sparse metrics are calculated for specific predicted locations based on observational data, with the specific calculation formulas as follows:

The Threat Score (TS):
\begin{equation}
\text{TS} = \frac{TP_1}{TP_1 + FN_1 + FP_1} \tag{A5}
\end{equation}

The F1 score:
\begin{equation}
\text{F1} = \frac{2TP_1}{2TP_1 + FN_1 + FP_1} \tag{A6}
\end{equation}

In Equations A1-A6, \( TP_i \) is the number of True Positives for class \( i \), \( TN_i \) is the number of True Negatives for class \( i \), \( FP_i \) is the number of False Positives for class \( i \), and \( FN_i \) is the number of False Negatives for class \( i \), where class 1 is marine fog and class 0 is background.

\subsubsection{Other Experimental Results} 

\begin{figure}[t] 
\centering 
\includegraphics[width=\textwidth]{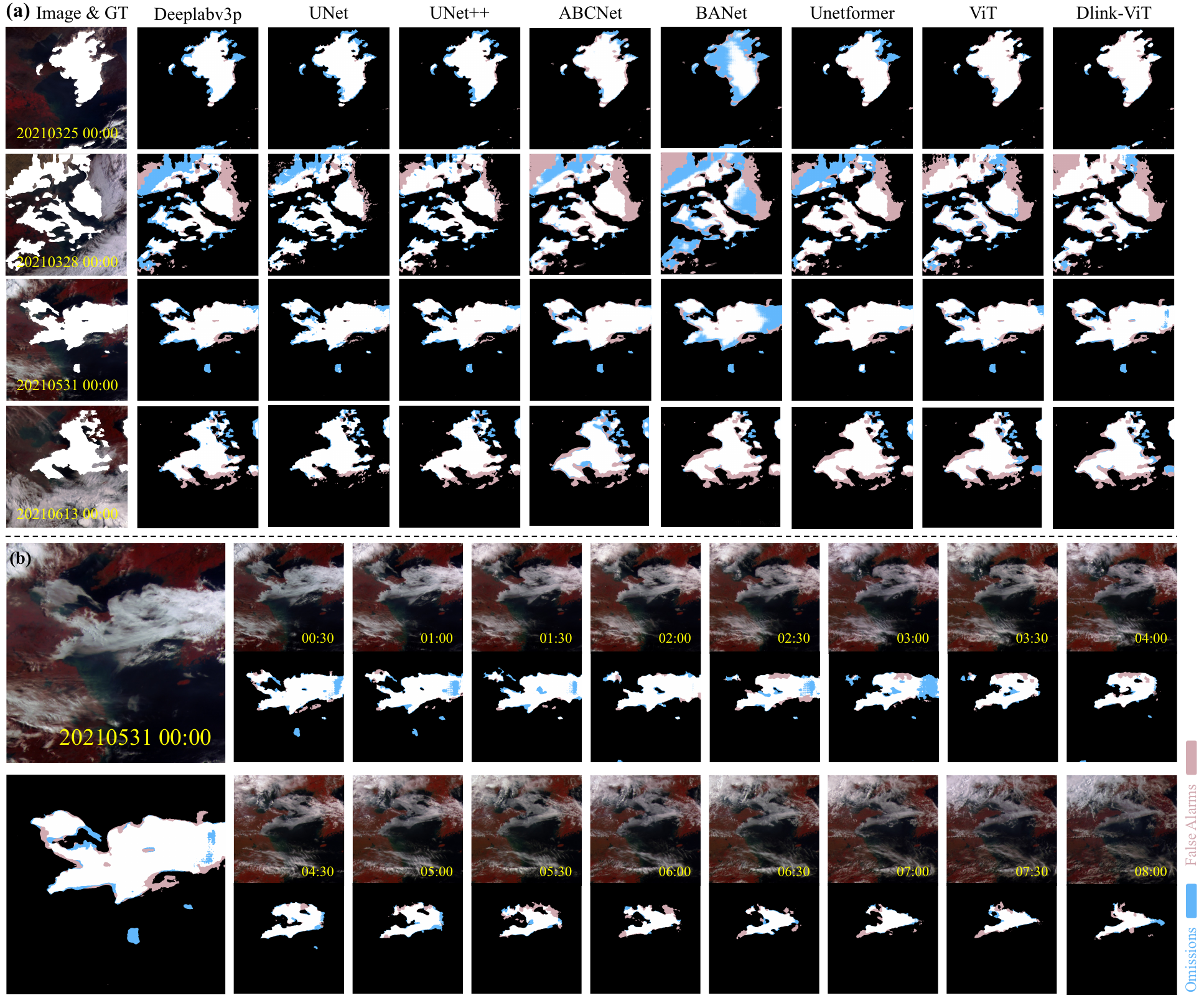} 
\caption{The qualitative visualization on Track A Static Marine Fog Detection includes (a) comparisons between different methods, and (b) an illustrative instance of uninterrupted marine fog surveillance on May 31, 2021 in Yellow and Bohai Sea based on FY4A satellite images. } 
\label{fig:Fig_tracka_fy4a} 
\end{figure}

Fig.~\ref{fig:Fig_tracka_fy4a} presents the qualitative experimental results of Track A Static Marine Fog Detection using FY4A sub-datasets. We compare the performance of multiple baseline methods at different timestamps. Additionally, visual inference on continuous satellite data with half of hour intervals is evaluated within a super data cube using Dlink-ViT-Base model.

Additionally, we provide more comprehensive experimental results to demonstrate the effectiveness and validity of the proposed super data cubes in M4Fog by utilizing the FY4A and H8/9 sub-datasets, along with four different typical baselines. Quantitative experimental results with dense metrics are presented in the Table.~\ref{tab:tracka_sst}.

\begin{table}[t]
\centering
\caption{The effectiveness of proposed super data cubes constructions in M4Fog on H8/9 and FY4A sub-datasets.}
\label{tab:tracka_sst}
\resizebox{\textwidth}{!}{%
  \begin{tabular}{c|c|cccc|cccc}
  \toprule
  \multirow{2}{*}{\textbf{Method}} & \multirow{2}{*}{\textbf{Settings}} & \multicolumn{4}{c|}{\textbf{H8/9 Sats. Dense-metrics}} & \multicolumn{4}{c}{\textbf{FY4A Sats. Dense-metrics}} \\
  & & \textbf{CSI} $\uparrow$ & \textbf{mIou} $\uparrow$ & \textbf{Acc} $\uparrow$ & \textbf{mAcc} $\uparrow$ & \textbf{CSI} $\uparrow$ & \textbf{mIou} $\uparrow$ & \textbf{Acc} $\uparrow$ & \textbf{mAcc} $\uparrow$ \\
  
  \midrule
  UNet            & +Cons. & 55.99 \tiny{\textcolor{red}{+1.51}}   & 76.96 \tiny{\textcolor{red}{+0.15}} & 70.74 \tiny{\textcolor{red}{+0.14}} & 84.88 \tiny{\textcolor{red}{+0.14}}  & 45.92 \tiny{\textcolor{red}{+0.15}} & 71.57 \tiny{\textcolor{red}{+0.28}} & 64.80 \tiny{\textcolor{green}{-8.62}} & 81.64 \tiny{\textcolor{green}{-3.95}} \\
  UNet++          & +Cons. & 55.74 \tiny{\textcolor{red}{+0.03}}   & 76.75 \tiny{\textcolor{green}{-0.03}} & 76.01 \tiny{\textcolor{red}{+2.69}} & 87.32 \tiny{\textcolor{red}{+1.25}}  & 47.60 \tiny{\textcolor{red}{+1.81}} & 72.33 \tiny{\textcolor{red}{+1.03}} & 72.97 \tiny{\textcolor{green}{-0.59}} & 85.50 \tiny{\textcolor{green}{-0.15}} \\
  ViT-Base        & +Cons. & 58.87 \tiny{\textcolor{red}{+1.94}}   & 78.48 \tiny{\textcolor{red}{+1.34}} & 73.85 \tiny{\textcolor{red}{+1.34}} & 86.45 \tiny{\textcolor{red}{+0.70}}  & 53.85 \tiny{\textcolor{red}{+1.08}} & 75.76 \tiny{\textcolor{red}{+0.68}} & 74.19 \tiny{\textcolor{green}{-5.19}} & 86.40 \tiny{\textcolor{green}{-2.03}}\\
  Dlink-ViT-Base  & +Cons. & 60.09 \tiny{\textcolor{green}{-0.16}} & 79.14 \tiny{\textcolor{green}{-0.10}} & 73.59 \tiny{\textcolor{red}{+1.67}} & 86.37 \tiny{\textcolor{red}{+0.77}}  & 55.79 \tiny{\textcolor{green}{-1.14}} & 76.73 \tiny{\textcolor{green}{-0.65}} & 79.92 \tiny{\textcolor{red}{+1.62}} & 89.16 \tiny{\textcolor{red}{+0.70}}\\
  
  \midrule
  UNet            & +SST  & 56.46 \tiny{\textcolor{red}{+1.98}}   & 77.22 \tiny{\textcolor{red}{+1.07}} & 56.46 \tiny{\textcolor{red}{+1.98}} & 84.92 \tiny{\textcolor{red}{+0.18}} & 46.17 \tiny{\textcolor{red}{+0.04}} & 71.72 \tiny{\textcolor{red}{+0.43}} & 64.12 \tiny{\textcolor{green}{-9.30}} & 81.34 \tiny{\textcolor{green}{-4.25}} \\
  UNet++          & +SST  & 55.70 \tiny{\textcolor{green}{-0.01}} & 76.73 \tiny{\textcolor{green}{-0.05}} & 76.30 \tiny{\textcolor{red}{+2.98}} & 87.46 \tiny{\textcolor{red}{+1.39}} & 47.58 \tiny{\textcolor{red}{+1.79}} & 72.32 \tiny{\textcolor{red}{+1.02}} & 72.79 \tiny{\textcolor{green}{-0.59}} & 85.41 \tiny{\textcolor{green}{-0.17}} \\
  ViT-Base        & +SST  & 59.00 \tiny{\textcolor{red}{+1.97}}   & 78.54 \tiny{\textcolor{red}{+1.04}} & 74.56 \tiny{\textcolor{red}{+2.05}} & 86.79 \tiny{\textcolor{red}{+1.04}} & 54.86 \tiny{\textcolor{red}{+1.91}} & 76.30 \tiny{\textcolor{red}{+1.22}} & 74.69 \tiny{\textcolor{green}{-4.69}} & 86.68 \tiny{\textcolor{green}{-2.08}} \\
  Dlink-ViT-Base  & +SST  & 60.23 \tiny{\textcolor{green}{-0.02}} & 79.19\tiny{\textcolor{green}{-0.05}} & 75.76 \tiny{\textcolor{red}{+3.84}} & 87.40 \tiny{\textcolor{red}{+1.80}} & 56.95 \tiny{\textcolor{red}{+0.02}} & 76.73 \tiny{\textcolor{green}{-0.35}} & 79.92 \tiny{\textcolor{red}{+1.62}} & 87.40 \tiny{\textcolor{green}{-1.06}} \\
  
  \midrule
  UNet            & +Cons.+SST & 55.89 \tiny{\textcolor{red}{+1.41}}  & 76.89 \tiny{\textcolor{red}{+0.74}} & 72.28 \tiny{\textcolor{red}{+1.68}} & 85.59 \tiny{\textcolor{red}{+0.85}} & 46.59 \tiny{\textcolor{red}{+0.82}} & 71.84 \tiny{\textcolor{red}{+0.55}} & 69.27 \tiny{\textcolor{green}{-4.15}} & 83.74 \tiny{\textcolor{green}{-1.85}} \\
  UNet++          & +Cons.+SST &55.56 \tiny{\textcolor{green}{-0.05}} & 76.69 \tiny{\textcolor{green}{-0.09}}& 73.55 \tiny{\textcolor{red}{+0.23}} & 86.17 \tiny{\textcolor{red}{+0.10}} & 47.40 \tiny{\textcolor{red}{+1.61}} & 72.16 \tiny{\textcolor{red}{+0.86}} & 75.67 \tiny{\textcolor{red}{+2.29}} & 86.73 \tiny{\textcolor{red}{+1.15}} \\
  ViT-Base        & +Cons.+SST &59.26 \tiny{\textcolor{red}{+2.23}}   & 78.66 \tiny{\textcolor{red}{+1.16}} & 75.62 \tiny{\textcolor{red}{+3.11}} & 87.29 \tiny{\textcolor{red}{+1.54}} & 55.23 \tiny{\textcolor{red}{+2.46}} & 76.44 \tiny{\textcolor{red}{+1.36}} & 78.95 \tiny{\textcolor{green}{-0.58}} & 88.68 \tiny{\textcolor{green}{-0.08}} \\
  Dlink-ViT-Base  & +Cons.+SST &60.81 \tiny{\textcolor{red}{+0.56}}   & 79.47 \tiny{\textcolor{red}{+0.25}} & 77.96 \tiny{\textcolor{red}{+6.04}} & 88.45 \tiny{\textcolor{red}{+2.85}} & 58.73 \tiny{\textcolor{red}{+1.80}} & 78.36 \tiny{\textcolor{red}{+0.98}} & 77.65 \tiny{\textcolor{green}{-0.65}} & 88.23 \tiny{\textcolor{green}{-0.23}} \\
  \bottomrule
  \end{tabular}%
}
\end{table}

\section{More Details for Track B Dynamic marine fog detection}
\label{sec:trackb_details}

Due to space limitations, only a portion of the dynamic marine fog detection results are presented in the main text. Here, the results of all 6 baseline experiments and 4 evaluation metrics are provided in Table.~\ref{tab:trackb_details}. We adapted the number of input and output heads to accommodate multi-temporal data. The results show that among the three baseline methods utilizing temporal information employed in this paper, the Dual-ways is the best, providing consistently improved model performance.

\begin{table}[ht]
\centering
\caption{The other detailed quantitative results with dense metrics for Dynamic Marine Fog Detection based on H8/9 sub-dataset when N=1.}
\label{tab:trackb_details}
\resizebox{\textwidth}{!}{%
  \begin{tabular}{c|c|cccc}
  \toprule
  \multirow{2}{*}{\textbf{Method}} & \multirow{2}{*}{\textbf{Settings}} & \multicolumn{4}{c}{\textbf{H8/9 Sats. Dense-metrics}}  \\
  & & \textbf{CSI} $\uparrow$ & \textbf{mIou} $\uparrow$ & \textbf{Acc} $\uparrow$ & \textbf{mAcc} $\uparrow$  \\
  
  \midrule
  UNet           & - (1 input)  & 54.17 & 76.02 & 69.37 & 84.17\\
  Temporal-UNet  & - (2 inputs)  & 53.86 \tiny{\textcolor{green}{-0.31}} & 75.89 \tiny{\textcolor{green}{-0.13}} & 71.70 \tiny{\textcolor{red}{+2.33}} & 85.29 \tiny{\textcolor{red}{+1.12}}\\
  UNet           & Dual-ways (2 inputs + 2 outputs)  & 55.75 \tiny{\textcolor{red}{+1.58}} & 76.85 \tiny{\textcolor{red}{+0.83}} & 71.02 \tiny{\textcolor{red}{+1.65}} & 85.01 \tiny{\textcolor{red}{+0.84}}\\
  MoANet (UNet-based) & +Motion (1 input + motion)  & 54.58 \tiny{\textcolor{red}{+0.14}} & 76.28 \tiny{\textcolor{red}{+0.26}} & 68.21 \tiny{\textcolor{green}{-1.16}} & 83.66 \tiny{\textcolor{green}{-0.51}}\\

  \midrule
  Dlink-ViT-Base & - (1 input)  & 60.48 & 79.39 & 71.91 & 85.36 \\
  Temporal-Dlink-ViT-Base & - (2 inputs)  & 59.65 \tiny{\textcolor{green}{-0.83}} & 78.87 \tiny{\textcolor{green}{-0.76}} & 77.25 \tiny{\textcolor{red}{+5.61}} & 88.08 \tiny{\textcolor{red}{+2.72}} \\
  Dlink-ViT-Base & Dual-ways (2 inputs + 2 outputs)  & 61.90 \tiny{\textcolor{red}{+0.42}} & 80.12 \tiny{\textcolor{red}{+0.73}} & 75.81 \tiny{\textcolor{red}{+3.90}} & 87.50 \tiny{\textcolor{red}{+2.14}} \\
  MoANet (Dlink-ViT-based) & +Motion (1 input + motion) & 57.69 \tiny{\textcolor{green}{-2.79}} & 77.83 \tiny{\textcolor{green}{-1.56}} & 75.66 \tiny{\textcolor{red}{+3.75}} & 87.26 \tiny{\textcolor{red}{+1.90}}\\

  \bottomrule
  \end{tabular}%
}
\end{table}

\section{More Details for Track C Spatio-temporal predictions}
\label{sec:trackc_details}

\subsection{Setup}

The data composition of each sub-dataset in Track C Spatio-temporal Prediction of cloud images is shown in Table.~\ref{tab:trackc_subdatasets} based on different satellites. Besides, we provide 736 samples of new maritime regions that have never been included in the training set as an additional test set for the GOES-16 dataset. This allows for a more comprehensive evaluation of the prediction model's performance.

\begin{table}[ht]
\centering
\caption{The details of sub-datasets constructed for Track C Spatio-Temporal Prediction for cloud imageries based on different satellites.}
\label{tab:trackc_subdatasets}
\tiny
\resizebox{\textwidth}{!}{%
\begin{tabular}{c|c|cc|c|ccc}
    \toprule
    \multirow{2}{*}{\textbf{Sub-datasets}} & \multirow{2}{*}{\textbf{Data Size}} & \multicolumn{2}{c|}{\textbf{Temporal Range}} & \multirow{2}{*}{\textbf{Regional Information}} & \multicolumn{3}{c}{\textbf{The Number of samples}} \\
    & & Range & Split & & Total & Training & Test \\
    
    \midrule
    \multirow{2}{*}{\textbf{Himawari-8/9}} & (8, 1024, 1024, 3) & 2017-2022 & 2021 & Specific (Yellow/Bohai Seas) & 2,512 & 1,840 & 672 \\
     & (8, 1024, 1024, 3) & 2017-2022 & 2021 & Multi-regional (2 areas) & 4,944  & 3,600 & 1,344 \\ 
     
    \textbf{FengYun-4A}   & (8, 1024, 1024, 3) & 2018-2021 & 2021 & Multi-regional (3 areas) & 3,931  & 2,878  & 1,053  \\
    \textbf{MeteoSat-11}  & (8, 1024, 1024, 3) & 2020-2022 & 2022 & Multi-regional (6 areas) & 2,832  & 2,016  & 816 \\
    \textbf{GOES-16}      & (8, 1024, 1024, 3) & 2020-2024 & 2024 & Multi-regional (6 areas) & 5,713  & 3,395  & 1,582+736 \\

    \bottomrule
\end{tabular}%
}
\end{table}

\subsection{Evaluation Metrics}
\subsubsection{Error Metrics}
The  Mean Absolute Error (MAE):
\begin{equation}
\text{MAE} = \frac{1}{T} \sum_{t=1}^T \sum_{c=1}^C \sum_{h=1}^H \sum_{w=1}^W |y_{t,c,h,w} - \hat{y}_{t,c,h,w}| \tag{A8}
\end{equation}

The  Mean Squared Error (MSE):
\begin{equation}
\text{MSE} = \frac{1}{T} \sum_{t=1}^T \sum_{c=1}^C \sum_{h=1}^H \sum_{w=1}^W (y_{t,c,h,w} - \hat{y}_{t,c,h,w})^2 \tag{A7}
\end{equation}

In Equations A7 and A8, $T,C,H$ and $W$ represent time, channel, height, and width, respectively; $y$ represents the prediction, and $\hat{y}$ represents the ground truth.

\subsubsection{Similarity Metrics}
The Peak Signal-to Noise Ratio (PSNR):
\begin{equation}
\text{PSNR} = 10 \cdot \log_{10} \left(\frac{\text{MAX}^2}{\text{MSE}}\right) \tag{A9}
\end{equation}

The Structural Similarity Index Measure (SSIM):
\begin{equation}
\text{SSIM} = \frac{(2 \mu_y \mu_{\hat{y}} + C_1)(2 \sigma_{y\hat{y}} + C_2)}{(\mu_y^2 + \mu_{\hat{y}}^2 + C_1)(\sigma_y^2 + \sigma_{\hat{y}}^2 + C_2)} \tag{A10}
\end{equation}

In Equation A9, \( \text{MAX} \) is the maximum pixel value of the prediction $y$, and \( \text{MSE} \) is the Mean Squared Error between $y$ and $\hat{y}$.
In Equation A10, \( \mu_y \) and \( \mu_{\hat{y}} \) are the mean values of $y$ and $\hat{y}$, \( \sigma_y^2 \) and \( \sigma_{\hat{y}}^2 \) are the variances of $y$ and $\hat{y}$, \( \sigma_{y\hat{y}} \) is the covariance of $y$ and $\hat{y}$. \( C_1 \) and \( C_2 \) are small constants used to stabilize the division.

\subsection{Other Experimental Results}
\begin{figure}[t] 
\centering 
\includegraphics[width=\textwidth]{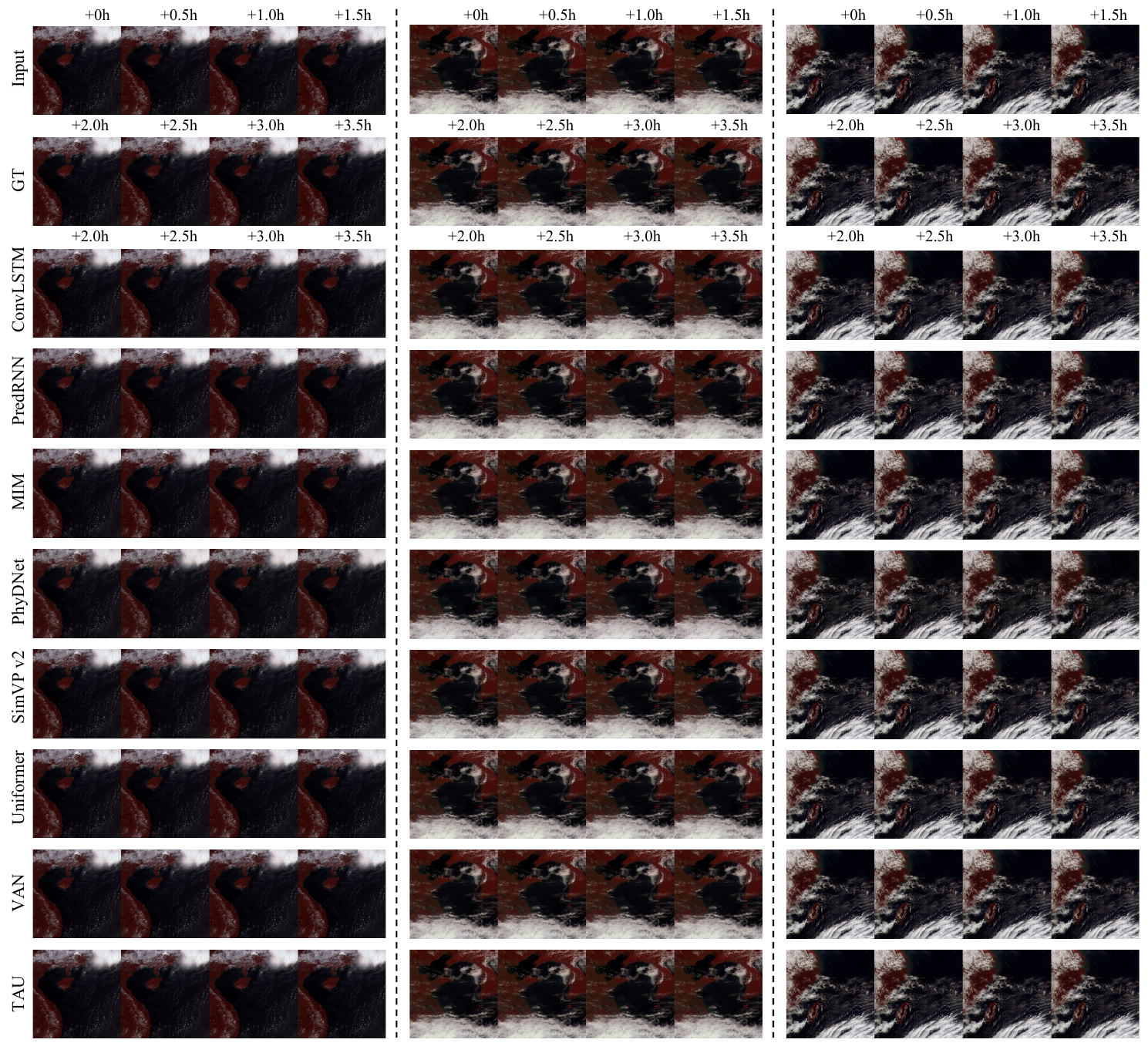} 
\caption{Visualization results of different Spatio-temporal Prediction for cloud images based on FY4A satellite data in Track C methods for different maritime regions, covering the Yellow Sea and Bohai Sea, the South China Sea and the East China Sea.} 
\label{fig:Fig_trackc_fy4a} 
\end{figure}

\begin{figure}[t] 
\centering 
\includegraphics[width=\textwidth]{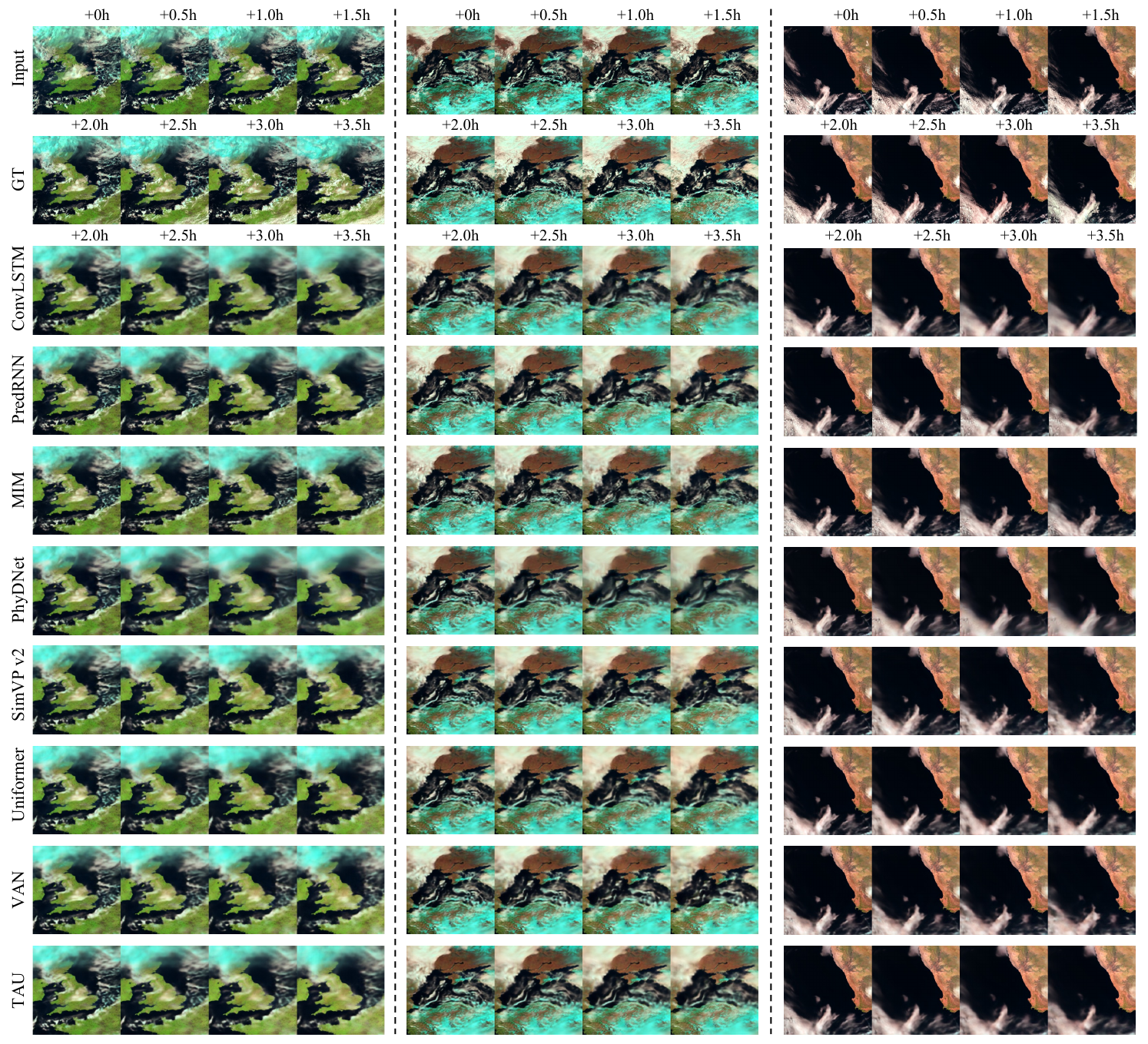} 
\caption{Visualization results of different Spatio-temporal Prediction for cloud images based on MSG (MeteoSat11) satellite data in Track C methods for different maritime regions, covering North Sea, Mediterranean, and the near Namibia seas.} 
\label{fig:Fig_trackc_meteo} 
\end{figure}

Table.~\ref{tab:trackc_fy4a} and \ref{tab:trackc_Meteo11} present the benchmark results of mainstream spatio-temporal prediction methods on the FY4A multi-region sub-dataset and the MeteoSat11 multi-region sub-dataset, respectively. The visualization results are shown in Fig.~\ref{fig:Fig_trackc_fy4a} and \ref{fig:Fig_trackc_meteo}. It can be observed that the FY4A satellite, due to the lack of a green visible light band, produces synthesized images with relatively uniform colors, making them easier to predict. As a result, the performance of all models improves significantly. Recurrent-free models are sufficient to capture the change patterns and avoid error accumulation, thus generally outperforming recurrent models. In contrast, the MeteoSat11 satellite images have more varied colors, making them more difficult to predict, resulting in noticeably lower performance metrics.

\begin{table}[h]
\centering
\caption{The performance on Track C Spatio-temporal prediction using continuous FY4A satellite data collected from super data cubes based on multi-areas.}
\label{tab:trackc_fy4a}
\tiny 
\resizebox{\textwidth}{!}{%
\begin{tabular}{c|cc|cccc}
    \toprule
    & \multicolumn{2}{c|}{\textbf{Method}} & \textbf{MSE} $\downarrow$ &  \textbf{MAE} $\downarrow$ & \textbf{SSIM} $\uparrow$ &  \textbf{PSNR} $\uparrow$ \\
    \midrule
    \multirow{8}{*}{\rotatebox{90}{\textbf{FY4 (4-4 Prediction)}}} & ConvLSTM [2015' NIPS]  & Recurrent-based & 45.77 & 2000.60  & 0.9831  & 36.61  \\
    & PredRNN   [2017' NIPS]   & Recurrent-based &  \underline{1.50}  & \underline{341.94}   & \underline{0.9995}  & \underline{51.88}  \\
    & MIM       [2019' CVPR]   & Recurrent-based &  35.54 & 1776.12  & 0.9868  & 37.65  \\
    & PhyDNet   [2020' CVPR]   & Recurrent-based & 130.81 & 3494.92  & 0.9491  & 32.07  \\
    & SimVP v2  [2022' Arxiv]  & Recurrent-free  &  4.89  & 618.09   & 0.9985  & 46.35  \\
    & Uniformer [2022' ICLR]   & Recurrent-free  &  \textbf{0.23}  & \textbf{118.57}   & \textbf{0.9999}  & \textbf{59.60} \\
    & VAN       [2022' Arxiv]  & Recurrent-free  &  4.95  & 617.28   & 0.9985  & 46.30  \\
    & TAU       [2023' CVPR]   & Recurrent-free  &  4.71  & 605.10   & 0.9986  & 46.53  \\
    \bottomrule
\end{tabular}%
}
\end{table}

\begin{table}[ht]
\centering
\caption{The performance on Track C Spatio-temporal prediction using continuous MSG (MeteoSat11) satellite data collected from super data cubes based on multi-areas.}
\label{tab:trackc_Meteo11}
\tiny 
\resizebox{\textwidth}{!}{%
\begin{tabular}{c|cc|cccc}
    \toprule
    & \multicolumn{2}{c|}{\textbf{Method}} & \textbf{MSE} $\downarrow$ &  \textbf{MAE} $\downarrow$ & \textbf{SSIM} $\uparrow$ &  \textbf{PSNR} $\uparrow$ \\
    \midrule
    \multirow{8}{*}{\rotatebox{90}{\textbf{MSG (4-4 Prediction)}}} & ConvLSTM [2015' NIPS]  & Recurrent-based & \textbf{2595.65} & 14692.31  & 0.6448  & \textbf{19.48}  \\
    & PredRNN   [2017' NIPS]   & Recurrent-based & 2733.22 & \textbf{14418.31} & \textbf{0.6709} & \underline{19.40} \\
    & MIM       [2019' CVPR]   & Recurrent-based & 2951.49 & 15203.85 & 0.6403 & 18.99 \\
    & PhyDNet   [2020' CVPR]   & Recurrent-based & \underline{2663.84} & 15053.23 & 0.6438 & 19.39 \\
    & SimVP v2  [2022' Arxiv]  & Recurrent-free  & 3138.52 & 15802.51 & 0.6309 & 18.75 \\
    & Uniformer [2022' ICLR]   & Recurrent-free  & 2894.25 & 15109.45 & 0.6467 & 19.09 \\
    & VAN       [2022' Arxiv]  & Recurrent-free  & 3098.12 & 15697.44 & 0.6321 & 18.79 \\
    & TAU       [2023' CVPR]   & Recurrent-free  & 2732.74 & \underline{14672.37} & \underline{0.6555} & 19.32 \\
    \bottomrule
\end{tabular}%
}
\end{table}

\end{document}